\documentclass[10pt,twocolumn,letterpaper]{article}

\usepackage[pagenumbers]{iccv} 
\usepackage{multirow} 
\usepackage{algorithm}
\usepackage{amsmath}
\usepackage{amssymb}
\usepackage{algpseudocode}
\definecolor{iccvblue}{rgb}{0.21,0.49,0.74}
\usepackage[pagebackref,breaklinks,colorlinks,allcolors=iccvblue]{hyperref}

\def\paperID{*****} 
\def\confName{ICCV}
\def\confYear{2025}

\title{Prompting Forgetting: Unlearning in GANs via Textual Guidance}

\author{Piyush Nagasubramaniam$^{1}$
\and
Neeraj Karamchandani$^{1}$
\and
Chen Wu$^{2}$
\and
Sencun Zhu$^{1}$\\
{$^1$The Pennsylvania State University \quad $^2$Meta}
}

\begin{document}
\maketitle
\begin{abstract}
State-of-the-art generative models exhibit powerful image-generation capabilities, introducing various ethical and legal challenges to service providers hosting these models. Consequently, Content Removal Techniques (CRTs) have emerged as a growing area of research to control outputs without full-scale retraining. Recent work has explored the use of Machine Unlearning in generative models to address content removal. However, the focus of such research has been on diffusion models, and unlearning in Generative Adversarial Networks (GANs) has remained largely unexplored. We address this gap by proposing Text-to-Unlearn, a novel framework that selectively unlearns concepts from pre-trained GANs using only text prompts, enabling feature unlearning, identity unlearning, and fine-grained tasks like expression and multi-attribute removal in models trained on human faces. Leveraging natural language descriptions, our approach guides the unlearning process without requiring additional datasets or supervised fine-tuning, offering a scalable and efficient solution. To evaluate its effectiveness, we introduce an automatic unlearning assessment method adapted from state-of-the-art image-text alignment metrics, providing a comprehensive analysis of the unlearning methodology. To our knowledge, Text-to-Unlearn is the first cross-modal unlearning framework for GANs, representing a flexible and efficient advancement in managing generative model behavior.
\end{abstract}    
\section{Introduction}
\label{sec:intro}
Generative image models, 
popularized by Generative Adversarial Networks (GANs)~\cite{goodfellowgan} and 
diffusion models~\cite{stablediffusionxl, ddpm}, can synthesize highly detailed, photorealistic images. As these models continue to advance, there is an increasing need for mechanisms that enable selective removal of learned concepts, ensuring fine-grained control over model outputs without requiring expensive, full-scale retraining. The ability to precisely remove unwanted concepts is critical for applications ranging from artistic integrity preservation to ethical AI development. Content removal methods from generative image models can be broadly categorized into \textit{filtering} and \textit{unlearning}-based strategies. Filtering strategies operate post-generation and use trained classifiers or rules to identify unwanted outputs without modifying the original model weights. They are lightweight and suitable in dynamic settings (\eg, identifying NSFW content from a prompt). Unlearning-based strategies involve finetuning the model to address the root cause. They are suitable long-term solutions and are particularly relevant to compliance issues when service providers cannot rely on filtering strategies that may fail to identify unwanted content. 

While recent research has explored unlearning methods in diffusion models, GANs introduce distinct challenges that require new approaches. Unlike diffusion models, which naturally integrate text-based conditioning for image generation and modification, GANs lack direct textual control, making feature-level unlearning significantly more complex. However, GANs remain relevant due to their unique advantages: (i) they generate images in a single forward pass, providing significant speed advantages over diffusion models, (ii) they are more resource efficient, and (iii) their latent spaces allow for fine-grained attribute control~\cite{stylespace, stylemc, bermano2022}.

Existing work on unlearning for GANs has remained largely limited in scope, often focusing on simplistic attribute erasure without addressing multi-attribute interactions or evaluating unlearning effectiveness in a systematic manner. To address these limitations, we propose Text-to-Unlearn: A Cross-Modal Approach to Unlearning in GANs. Our approach uses natural language descriptions to guide the unlearning process, allowing for targeted concept removal without the need for additional datasets. Our key contributions include:
\begin{itemize} 
\item A novel unlearning framework that removes learned concepts in GANs using \emph{only} a text prompt, eliminating the need for additional data collection or supervised fine-tuning.
\item An extension of generative unlearning in GANs to complex, fine-grained tasks, including expression unlearning, multi-attribute unlearning, and disentangled feature removal.
\item A new quantitative evaluation metric, \emph{degree of unlearning} ($\gamma$), designed to measure unlearning performance using state-of-the-art Vision-Language Models (VLMs).
\end{itemize}

\section{Related Work}
\subsection{Machine Unlearning}
Machine Unlearning~\cite{sisa} was originally developed as a solution to support the \emph{right-to-be-forgotten}, which is a requirement of privacy regulations like GDPR~\cite{gdpr} and CCPA~\cite{ccpa}. The goal was to erase the influence of selected data samples without incurring the cost of retraining the model from scratch. Beyond user privacy, unlearning has been adopted for correcting biases and confusion in deep learning models~\cite{scrub}. In such cases, the error for a particular class (e.g., in classification) is maximized. More recently, unlearning has also been leveraged to eliminate backdoor attacks in deep learning models~\cite{wu2023unlearning, corrective}. One might observe the recurring theme that Machine Unlearning has traditionally been studied in a supervised setting. As such, there is much to explore regarding how unlearning can be adapted to solve problems pertaining to generative models.

\subsection{Generative Image Models and Unlearning}
To address the potential misuse of generative image models (\eg, StyleGAN2~\cite{stylegan2}, Stable Diffusion~\cite{stablediffusionxl}, DALL-E2~\cite{dalle2}, \etc), recent work has explored the idea of generative unlearning. For example, ~\citeauthor{guide} proposed GUIDE~\cite{guide} for identity unlearning in GANs using a reference image. Using GUIDE, individual identities can be unlearned even if the identity has not been seen during training. GUIDE makes use of a latent target unlearning method and an adjacency aware loss to ensure that all points in the latent space corresponding to an identity map to a different identity after the unlearning process while preserving the overall utility of the model. In this case, the problem is to effectively map neighboring points in the latent space without damaging the pre-trained GAN's performance; however, we consider a relatively broad problem in which text prompts can describe several types of features that are not necessarily close to each other in the latent space. 

~\citeauthor{Moon_2024} explored feature unlearning in VAEs and GANs (\eg, unlearning features like ``bangs"). They rely on curated datasets that are used to finetune the model as part of the unlearning process. The features of unlearning are based on annotations provided by datasets (\eg, CelebA) or frameworks like Morpho-MNIST~\cite{morpho_mnist} which can measure the extent of learning for features in the MNIST dataset. We consider the unlearning problem with fewer assumptions and study its application for models pre-trained on large high resolution datasets like Flickr-Faces HQ $1024 \times 1024$ (FFHQ) where such data curation may be infeasible or inaccessible due to privacy reasons.

Other than the aforementioned research, there has not been much work related to generative unlearning for GANs. However, unlearning and concept erasure has been studied within the context of diffusion models. Safe Latent Diffusion (SLD)~\cite{sld} is a method to mitigate the generation of content at inference time without the need for any additional finetuning of the diffusion model. Recent work like MACE~\cite{mace}, Erased Stable Diffusion (ESD)~\cite{esd}, and Forget-Me-Not~\cite{forgetmenot} propose various methods to finetune diffusion models and perform concept erasure. Our proposed method broadly falls into this category of methods, which uses finetuning to erase knowledge from the model.

\section{Problem Statement}
As mentioned earlier, unlike existing unlearning research, class labels may not always be available or even applicable in the context of generative models. Also, collecting images even for the purpose of unlearning may be challenging due to privacy regulations like GDPR and CCPA. 
Thus, the motivation for the methods discussed in this paper arises from the following question: \emph{Can we flexibly unlearn concepts from a GAN using only text prompts?}

\begin{figure}[ht]
    \centering
    \includegraphics[scale=0.5]{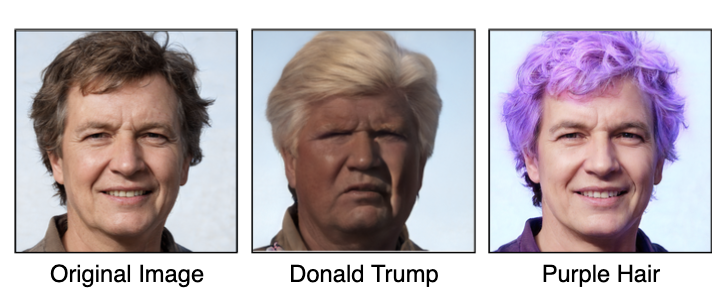}
    \caption{Examples of StyleCLIP manipulations of an image. The driving text for the edit is listed below each image.}
    \label{fig:styleclip_example}
\end{figure}

As shown in StyleCLIP~\cite{styleclip}, text prompts can be used to drive image manipulation in the latent space of the GAN to make either fine-grained edits or even incorporate features of popular individuals using the power of CLIP's~\cite{clip} joint embedding space. We show some relevant examples of StyleCLIP manipulations in Figure~\ref{fig:styleclip_example}.

\begin{figure*}[ht]
    \centering
    \includegraphics[width=\linewidth]{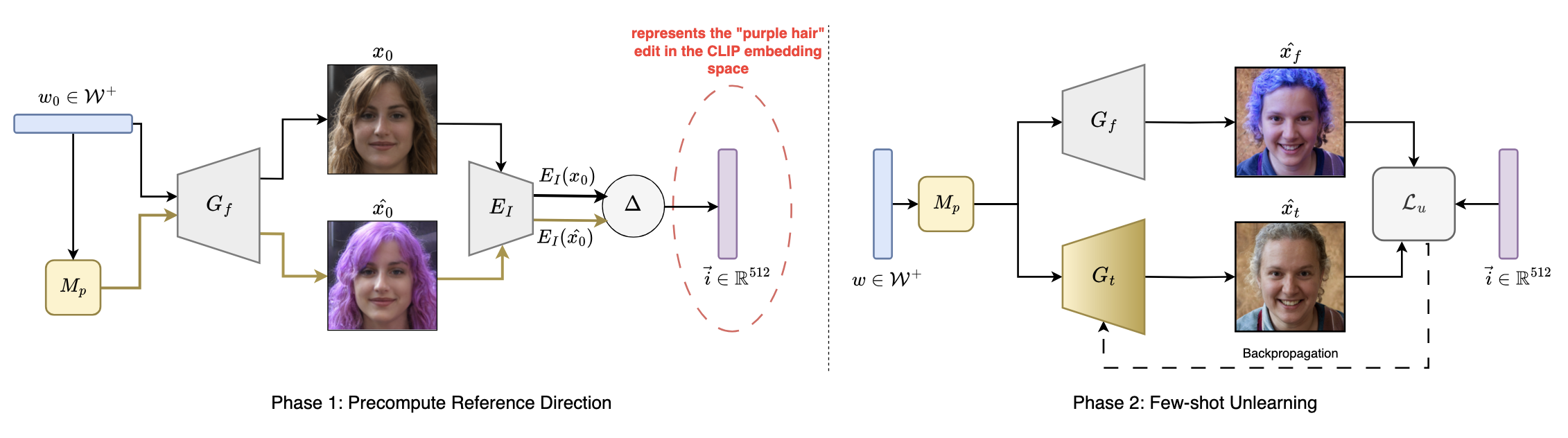}
    \caption{Overview of the Text-to-Unlearn framework for unlearning the feature ``purple hair" as an example. In the first phase, a reference direction to guide the unlearning is precomputed once. In the second phase, the precomputed reference direction is used to steer the trainable generator's synthesis network away from generating undesirable images.}
    \label{fig:unlearning_scheme}
\end{figure*}

Modern text-to-image generation models such as StyleGAN-T~\cite{stylegant}, Stable Diffusion, and DALL-E2 can generate images from textual descriptions that are only limited to the user's imagination. As such, we believe the ability to unlearn must be just as flexible as the image generation process. Thus, our framework is centered around using \emph{only} text prompts as a driver for the unlearning process to support unlearning at different levels of granularity (\eg, unlearning a hairstyle, hair color, identity, \etc) like those shown in Figure~\ref{fig:styleclip_example}. 

Now, we formalize the unlearning problem in GANs. We assume that the original training dataset is not available during the unlearning process to comply with the aforementioned privacy regulations. Furthermore, we do not require \emph{any} additional ``forget" samples to be collected by the training authority for the unlearning process. The only requirement is the model and a text prompt that describes the undesirable feature or concept to be erased. To our knowledge, this is the first work to explore unlearning in GANs with such limited assumptions and versatility.

Consider a pre-trained and trainable GAN Generator $G_{t}(\theta_0)$ where $\theta_0$ represents the initial model parameters. We wish to develop an unlearning strategy $\Lambda$ that unlearns a concept described by text prompt $p$. Formally, the result of the unlearning strategy is a target generator $G_{t}(\theta)$ where $\theta$ represents the updated model parameters:
\begin{equation}\label{eq:formulation}
    G_{t}(\theta) \triangleq \Lambda(G_{t}(\theta_0), p)
\end{equation}
With the updated parameters $\theta$, the generator should no longer be able to generate images with the undesirable feature or concept described by the text prompt $p$ while preserving the performance on other concepts. For example, if we wish to unlearn the feature ``purple hair", the unlearning strategy $\Lambda$ should not affect the ability of $G_{t}(\theta)$ to generate images with blonde hair. 

However, there are challenges to unlearning in GANs:
\begin{itemize}
    \item Erasing concepts from the GAN's latent space without affecting the overall image synthesis quality is difficult due to entanglement in the latent space, \ie, erasing one concept can easily affect the generation of several other features.
    \item Unlike diffusion models, GANs do not have textual inputs to generate samples for the unlearning process. Finding interpretable directions in the GAN's latent space for each dataset is intractable at scale.
    \item A key challenge specific to unlearning in GANs is the difficulty in measuring the extent to which unlearning is successful because it can be subjective.
\end{itemize}
\section{Methodology}
In this section, we first discuss the components of our framework (shown in Figure~\ref{fig:unlearning_scheme}) and then introduce one of our core contributions: \emph{directional unlearning}. Our methodology is motivated by the few-shot domain adaptation scheme in StyleGAN-NADA~\cite{stylegannada}.
\subsection{Overview}
Our framework consists of four key components: a latent mapper $M_p$ trained on a text prompt $p$ that describes the concept to be unlearned, a frozen copy of the generator $G_{f}$, a trainable generator $G_{t}$, and a pre-trained CLIP~\cite{clip} model.

\noindent \textbf{Latent Mapper ($M_p$).}
The latent mapper $M_p$ (as described in StyleCLIP~\cite{styleclip}) is a shallow neural network that maps latent codes within StyleGAN's $\mathcal{W^+}$ space, \ie, $M_p: \mathcal{W^+} \rightarrow \mathcal{W^+}$ and is used to edit images according to the prompt $p$. Suppose a point $w \in \mathcal{W^+}$ corresponds to an image of a man with black hair and the text prompt $p$ is ``purple hair". Then, the latent mapper can be used to compute $\hat{w} = w + M_p(w)$ such that $\hat{w}$ corresponds to an image of the same person with the only difference being purple hair. Simply put, we can use the latent mapper to edit any image according to a text description.

\noindent \textbf{Generators.} The trainable GAN generator $G_{t}$ will be finetuned using our unlearning strategy and will no longer produce images with the unlearned feature after the unlearning process is complete. $G_{f}$ is a copy of $G_{t}$ before unlearning and is used to generate images containing the feature to be unlearned.

\noindent \textbf{Pre-trained CLIP model ($E$).} We use a standard pre-trained CLIP model $E$ to capture images in a joint embedding space. We refer to CLIP's visual encoder as $E_I$. 

\subsection{Directional Unlearning}
The overall idea is to finetune $G_{t}$ using a few samples taken from its latent space so that it does not produce images containing the undesirable properties described by the text prompt. Since GANs are prone to mode collapse, $G_{f}$ and $M_p$ are used to help generate specific images, which are used to regularize the training with appropriate loss components. After the finetuning (unlearning) is complete, $G_{f}$ can be discarded. 

Our unlearning process is based on guiding $G_{t}$ along a direction in the CLIP embedding space derived from the text prompt $p$, so we deem our method as \emph{directional unlearning}. The process is split into two phases: (i) Phase 1: Precomputing a reference direction for unlearning, and (ii) Phase 2: Few-shot Unlearning.

\noindent \textbf{Phase 1.}
We choose a randomly sampled batch of latent codes $w_0 \in \mathcal{W^+}$ called the initial latent codes. The latent mapper uses the initial latent codes to compute $\hat{w_0} = w_0 + M_p(w_0)$. Then, the corresponding image batches for $w_0$ and $\hat{w_0}$ are given by $x_0$ and $\hat{x_0}$ (e.g., with purple hair):
\begin{equation}
    x_0 = G_{f}(w_0),~ \hat{x_0} = G_{f}(\hat{w_0})
\end{equation}
Once the pairs of image batches are computed, we compute a unit vector $\vec{i}$ (operation denoted by $\Delta$ in Figure~\ref{fig:unlearning_scheme}) representing the edit direction in the CLIP embedding space. Specifically,
\begin{equation}\label{eq:ref_direction}
    \vec{i} = \frac{E_I(\hat{x_0})-E_I(x_0)}{\| E_I(\hat{x_0})- E_I(x_0) \|_2}
\end{equation}
where $E_I(\cdot)$ represents the CLIP visual encoder, and $E_I(x_0)$ and $E_I(\hat{x_0})$ are the CLIP embeddings of the image batches $x_0$ and $\hat{x_0}$, respectively. 
Essentially, we capture the change of adding $p$ (e.g.,``purple hair") in the embedding space and later unlearn along this direction.

\noindent \textbf{Phase 2.} 
Now, we perform few-shot unlearning using the reference direction $\vec{i}$ (from Equation~\ref{eq:ref_direction}) from Phase 1. During each finetuning step, a batch of latent codes $w \in \mathcal{W^+}$ is sampled and passed through the latent mapper $M_p$ to generate latent codes $\hat{w} = w + M_p(w)$. The latent codes $\hat{w}$ will be provided as input to $G_{t}$ and $G_{f}$. The rationale for using a frozen generator is the same as StyleGAN-NADA~\cite{stylegannada}, \ie, to ensure that optimization favors solutions on the real image manifold. During the finetuning process, $G_{f}$ will constantly generate negative samples, \ie, images containing undesirable attributes described by $p$. However, $G_{t}$ will adapt to generate the same images without the undesirable attributes because the loss function $\mathcal{L}_u$ uses the precomputed reference direction $\vec{i}$ to guide the unlearning only along this direction. 

During each step, we use the adaptive layer selection method based on the global CLIP loss described in StyleGAN-NADA to prevent mode collapse by updating only a subset of $G_{t}$'s parameters. Furthermore, the weights of $G_{t}$'s mapping network are frozen and only the synthesis network is updated.

\noindent \textbf{Loss Function.}
First, we define the unlearning loss function $\mathcal{L}_u$ for \textit{feature unlearning}:
\begin{equation}\label{eq:loss1}
\begin{split}
    \mathcal{L}_u =~ &\mathcal{L}_{dir}(\hat{x_{t}}, \hat{x_{f}},\vec{i}) +\lambda_{lpips}\mathcal{L}_{lpips}(\hat{x_{t}}, \hat{x_{f}}) ~+ \\ & \lambda_{id}\mathcal{L}_{id}(\hat{x_{t}}, \hat{x_{f}})
\end{split}
\end{equation}

\begin{figure}
    \centering
    \includegraphics[scale=0.35]{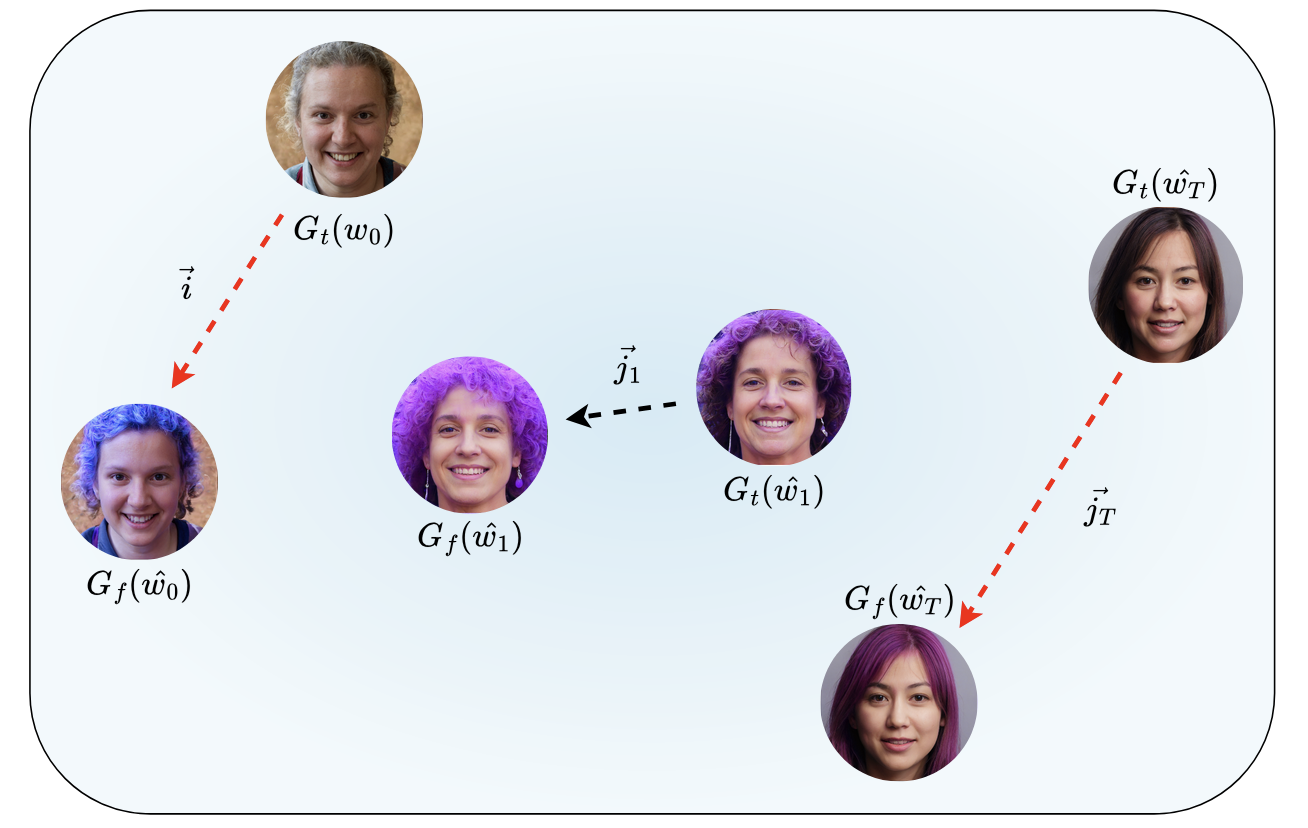}
    \caption{Examples of image embeddings in the CLIP space during the fine-tuning of $G_{t}$. $\vec{i}$ is the precomputed reference direction and, $\vec{j_1}$ and $\vec{j_T}$ are alignments during and at the end of training, respectively.}
    \label{fig:clip_space}
\end{figure}
\begin{figure*}
    \centering
    \begin{subfigure}[b]{0.3\linewidth}
        \includegraphics[width=\linewidth]{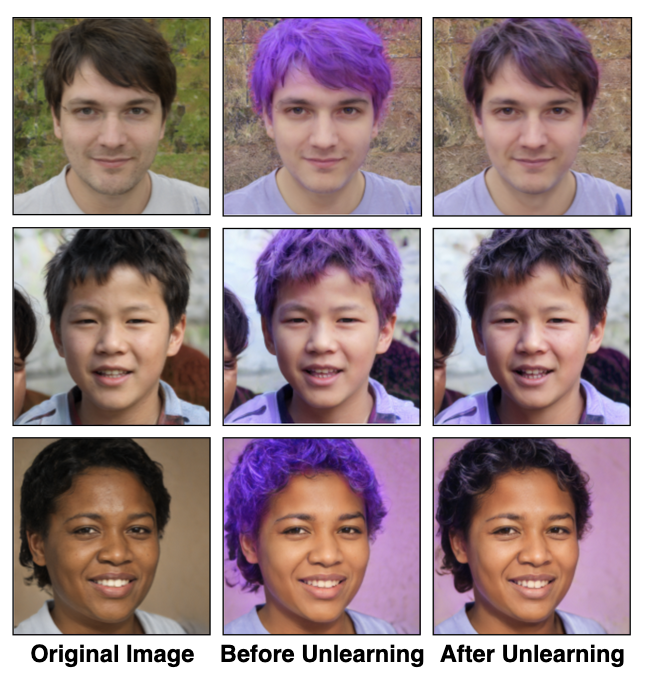} 
        \caption{Purple Hair}
    \end{subfigure}%
    \begin{subfigure}[b]{0.3\linewidth}
        \includegraphics[width=0.99\linewidth]{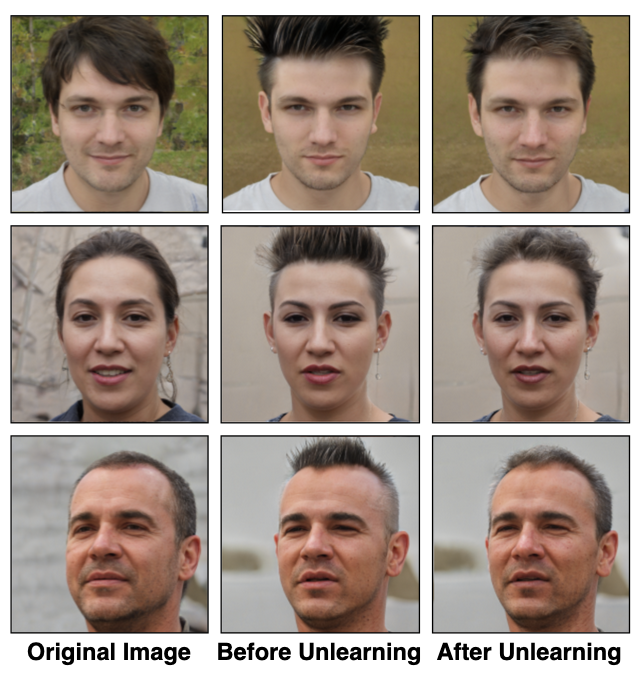} 
        \caption{Mohawk Hairstyle}
    \end{subfigure}%
    \begin{subfigure}[b]{0.3\linewidth}
        \includegraphics[width=0.98\linewidth]{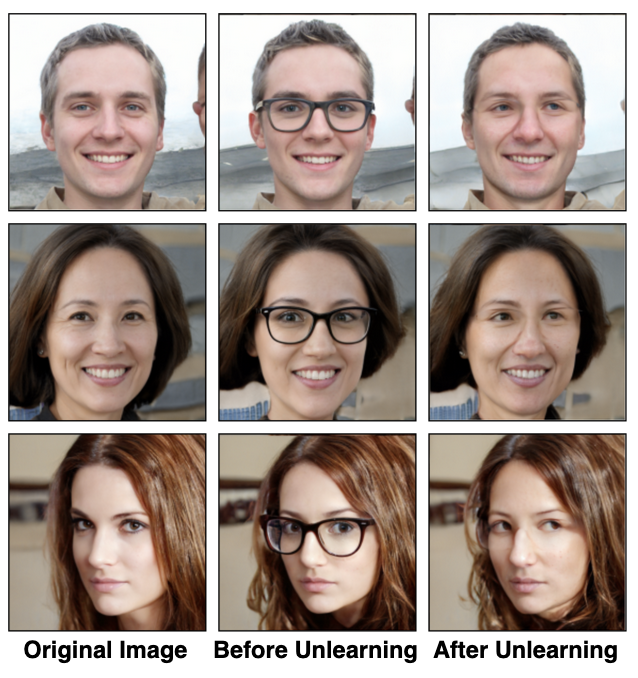} 
        \caption{Spectacles}
    \end{subfigure}%
    \caption{Qualitative comparison of generated images before and after unlearning features based on the text prompt (below each grid). The second column corresponds to a latent code that has an undesirable feature and the third column is the image synthesized from the same latent code after unlearning.}
    \label{fig:feat_unlearning}
\end{figure*}

Here, $\mathcal{L}_{dir}$ is the directional loss, $\mathcal{L}_{lpips}$ is the LPIPS loss for perceptual similarity~\cite{lpips}, and $\mathcal{L}_{id}$ is an ID loss based on the ArcFace facial recognition network~\cite{arcface, styleclip}. $\hat{x_{f}}$ and $\hat{x_{t}}$ are the images generated by $G_{f}$ and $G_{t}$, respectively. $\vec{i}$ is the precomputed reference direction from Equ.~\ref{eq:ref_direction}. 
The directional loss $\mathcal{L}_{dir}$ is the key component that guides the trainable generator $G_{t}$ away from synthesizing undesirable features. However, while unlearning the features, we need to preserve the usability of the latent space for downstream tasks like image manipulation and domain adaptation, and thus, we regularize the training process using ID loss and LPIPS loss. For example, while unlearning hair color or hairstyle, we would like to preserve the remaining features of the face. By using this method, latent mappers trained before unlearning can still be used to generate edits (except for prompts pertaining to the unlearned concept).

Suppose that $d_{cos}(\cdot)$ represents the cosine similarity function, then the directional loss is defined as:
\begin{equation}
\begin{split}
    &\mathcal{L}_{dir}(\hat{x_{t}},\hat{x_{f}},\vec{i}) = 1 - d_{cos}(\vec{i},\vec{j})\\
    &\vec{j} = \frac{E_I(\hat{x_{t}})-E_I(\hat{x_{f}})}{  \|E_I(\hat{x_{t}})-E_I(\hat{x_{f}})\|_2  }
\end{split}
\end{equation}

The unit vector $\vec{j}$ describes how the output images generated by $G_{t}$ and $G_{f}$ differ in the CLIP embedding space. During the unlearning process, we want $\vec{j}$ to be aligned with our precomputed reference direction $\vec{i}$, which does not change during training as described earlier. Clearly, minimizing $\mathcal{L}_{dir}$ rewards the alignment of $\vec{i}$ and $\vec{j}$, and this happens when the images generated by $G_{t}$ progressively exclude the undesired feature. To illustrate how the directional loss encourages the unlearning of a concept, consider a simplified example in Figure~\ref{fig:clip_space}. For the initial batch of latent codes $w_0$, a reference direction $\vec{i}$ is computed based on Equation~\ref{eq:ref_direction}. For the first batch $\hat{w_1}$ during training, observe that the output of $G_{t}$ still retains purple hair but it is less prominent. Consequently, $\vec{j_1}$ is not aligned with $\vec{i}$ at this time step. At the final time step $T$ of training, the vector $j_T$ perfectly aligns with the reference direction $\vec{i}$ since the image $G_{t}(\hat{w_T})$ does not contain any purple hair. In this example, the input to $G_{f}$ is always a latent code corresponding to an image with purple hair. The only way for $\vec{j}$ to align with $\vec{i}$ is if $G_{t}$ synthesizes images without purple hair by learning \emph{along} the vector $\vec{i}$.

In the case of \textit{identity unlearning}, we formulate a different unlearning loss $\mathcal{L}_{u,id}$ such that $G_{t}$ directs images toward the mean latent. Here, we only use the LPIPS loss to ensure optimization favors images from the original domain. Suppose the mean latent is given by $\overline{w} \in \mathcal{W^+}$ and the corresponding image is $\overline{x} = G_{t}(\overline{w})$, then the unlearning loss is given by:
\begin{equation}\label{eq:loss2}
    \mathcal{L}_{u,id} = \mathcal{L}_{dir}(\hat{x_{t}}, \overline{x},\vec{i}_{id}) + \mathcal{L}_{lpips}(\hat{x_{t}}, \overline{x})
\end{equation}
We define $\vec{i}_{id}$ as the precomputed reference direction for identity unlearning. Now, it is computed with respect to the mean latent instead of negative samples from the latent mapper as shown in Equation~\ref{eq:ref_dir_id}.
\begin{equation}\label{eq:ref_dir_id}
    \vec{i}_{id} = \frac{E_I(x_0)-E_I(\overline{x})}{\| E_I(x_0)-E_I(\overline{x}) \|_2},~ \vec{j}_{id} = \frac{E_I(\hat{x_{t}})-E_I(\overline{x})}{\| E_I(\hat{x_{t}})-E_I(\overline{x}) \|_2}
\end{equation}
Here, $x_0$ is the batch of images randomly sampled at the start of Phase 1. The underlying optimization problem remains the same as before and follows the same high-level idea depicted in Figure~\ref{fig:clip_space}.
Notably, the difference from StyleGAN-NADA is that we do not rely on source-target text pairs for the unlearning process. Unlike domain adaptation, the text phrases alone are unable to capture the fine-grained nature of unlearning. The latent mapper helps generate images, which can be used to implicitly capture the direction of the prompt $p$ in the CLIP embedding space for stable unlearning.
\begin{figure*}[ht]
    \centering
    \begin{subfigure}[b]{0.3\linewidth}
        \includegraphics[width=\linewidth]{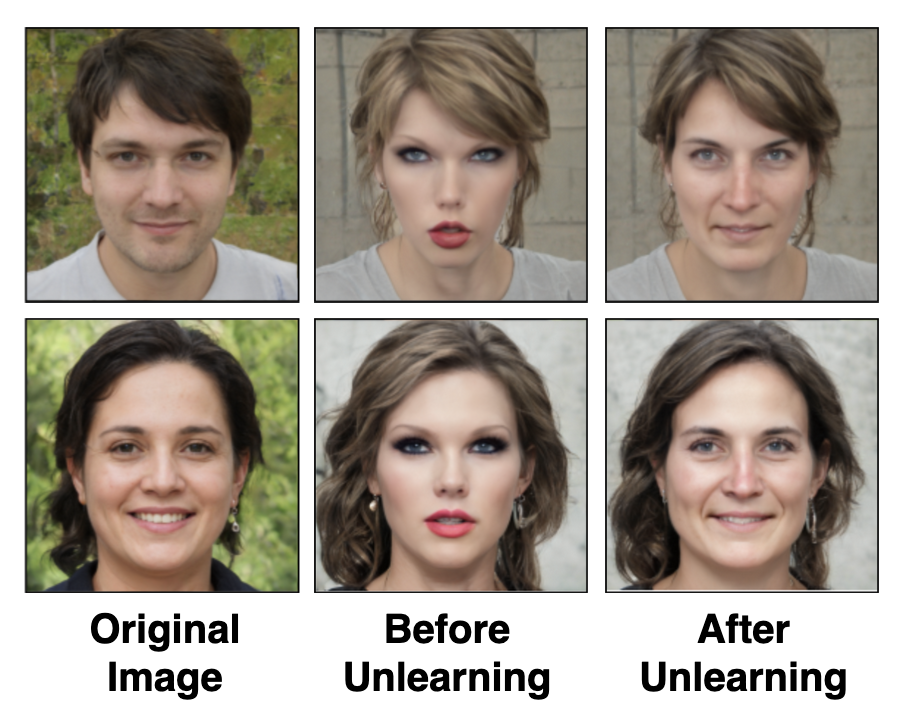} 
        \caption{Taylor Swift}
        \label{fig:taylor_subfig}
    \end{subfigure}%
    \begin{subfigure}[b]{0.3\linewidth}
        \includegraphics[width=0.99\linewidth]{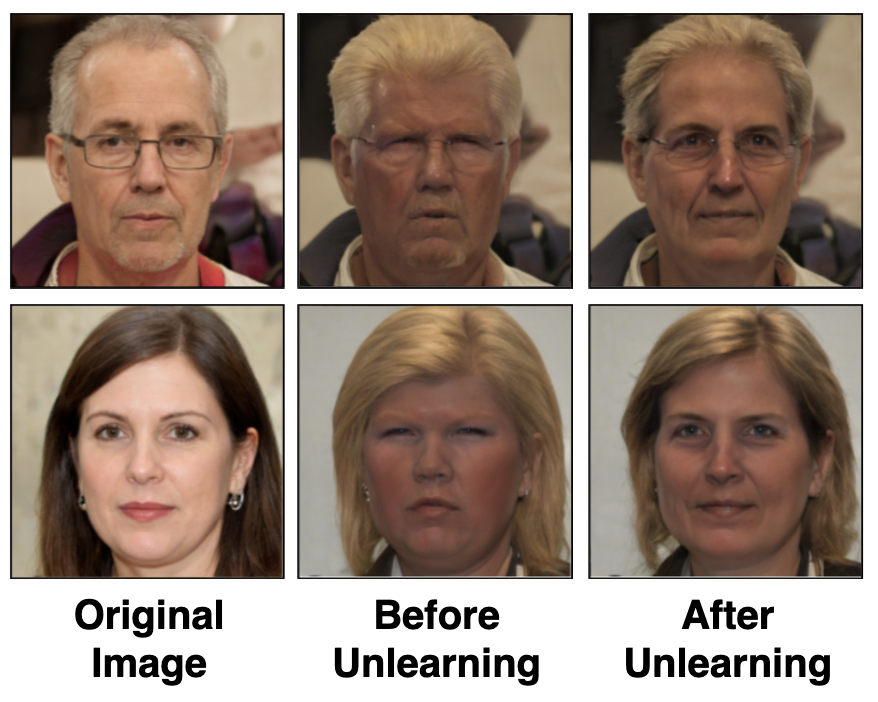} 
        \caption{Donald Trump}
    \end{subfigure}%
    \begin{subfigure}[b]{0.3\linewidth}
        \includegraphics[width=\linewidth]{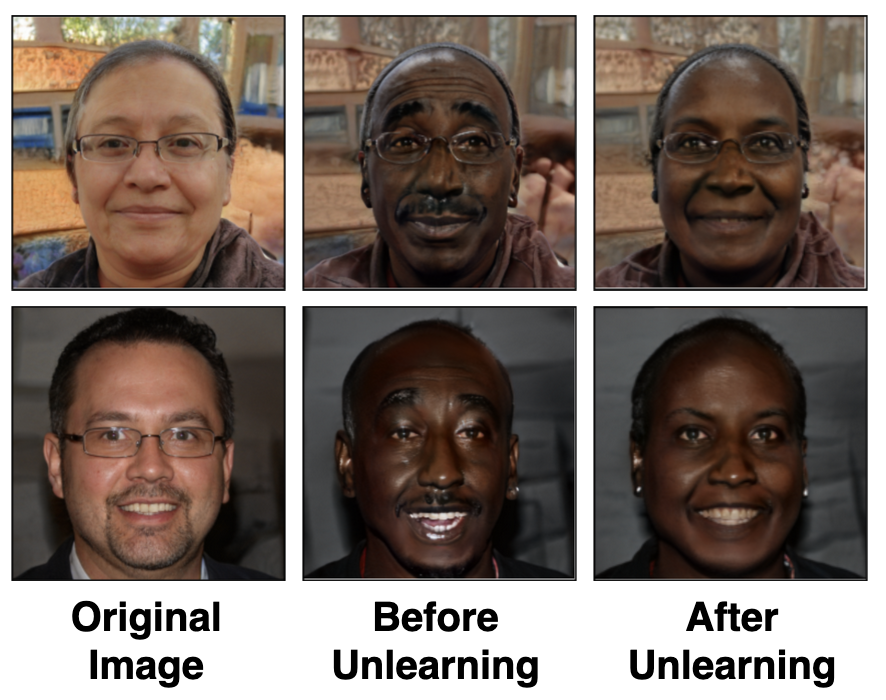} 
        \caption{Tupac}
    \end{subfigure}%
    \caption{Qualitative comparison of generated images before and after identity unlearning. The first column shows source samples from StyleGAN2. The second column shows images generated using the driving text (below each grid) on the source samples before unlearning. The third column shows the images for the same points after unlearning.}
    \label{fig:id_unlearning}
\end{figure*}
\section{Experiments}
\label{sec:experiments}
In this section, we discuss the results of our experiments for a variety of tasks including feature unlearning and identity unlearning.

\subsection{Experimental Setup}
We use StyleGAN2 pre-trained on the FFHQ dataset with an output resolution of $1024 \times 1024$ for our experiments. We do not explicitly use a separate dataset for the unlearning process. All samples needed for finetuning are sampled directly from the GAN's latent space. We include training details for the latent mappers and finetuning process in the Appendix.
\begin{figure}
    \centering
    \includegraphics[width=.8\columnwidth]{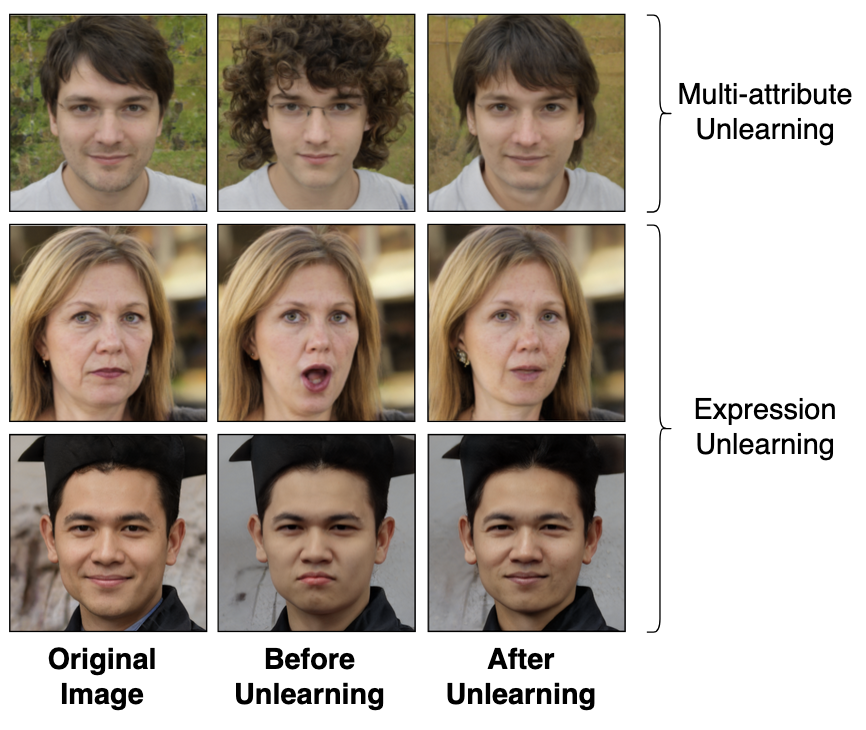}
    \caption{Examples of non-standard unlearning tasks including multi-attribute and expression unlearning. Top row: ``curly long hair", middle row: ``surprised", and bottom row: ``angry".}
    \label{fig:non_standard}
\end{figure}
\subsection{Qualitative Results}
\noindent \textbf{Feature Unlearning.}
We consider unlearning the following features of varying granularity: hair color, hairstyle, and accessories. The results of the GAN before and after unlearning are shown in Figure~\ref{fig:feat_unlearning}, and we see that for any chosen source image, the latent mapper can generate an edit with an undesirable feature. Using our text-guided unlearning scheme, the latent codes of images with undesirable features are now mapped to variations of the source image without those features. \\
\noindent \textbf{Identity Unlearning.}
Our Text-to-Unlearn framework is based on unlearning using only text prompts. As such, it is not within our scope to unlearn \textit{any} identity since accessing each identity from a text prompt is not possible. However, GAN manipulation frameworks like StyleCLIP~\cite{styleclip} and StyleGAN-NADA~\cite{stylegannada} can use driving text prompts like ``Beyonce" or ``Taylor Swift" to leverage CLIP's understanding of popular celebrities (presumably seen during pre-training) to incorporate their features. 
\begin{figure}
    \centering
    \includegraphics[width=0.8\columnwidth]{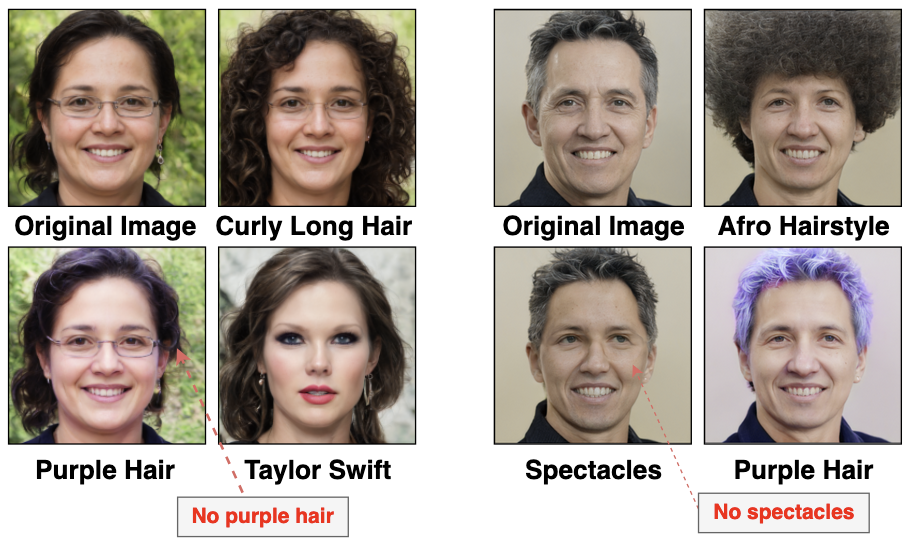}
    \caption{Example of using latent mappers to make edits after unlearning purple hair (left) and spectacles (right) from the GAN. The manipulation prompt is listed below each image.}
    \label{fig:manipulation}
\end{figure}

Thus, we consider the task of unlearning identities that are accessible through CLIP's text encoder. The results of unlearning identities using Equation~\ref{eq:loss2} are shown in Figure~\ref{fig:id_unlearning}.
Unlike feature unlearning, the images corresponding to the unlearned latent codes lack resemblance to the source images because we direct them toward the mean latent during training. The changes in hair color, hairstyle, \etc. are relatively fine-grained compared to identity manipulation and so, Equation~\ref{eq:loss2} is specifically designed to ensure the identity is erased instead of preserving it. We choose to direct the target latent toward the mean latent similar to GUIDE~\cite{guide} because the mean latent represents the average ``face" of the learned distribution, ensuring maximal stability during unlearning. 

\noindent \textbf{Non-Standard Unlearning Tasks.}
In addition to existing unlearning tasks like feature and identity unlearning, we leverage the disentangled $\mathcal{W^+}$ space to perform expression unlearning and multi-attribute unlearning. The key advantage of our text-to-unlearn method is the flexibility provided by text prompts. We can use the text prompts to unlearn multiple undesired features using a single text prompt. Similarly, we can also unlearn expressions from the model. The results for the unlearning prompts ``curly long hair", ``surprised", and ``angry" are shown in Figure~\ref{fig:non_standard}.
\begin{figure}[ht]
    \centering
    \begin{subfigure}[b]{.45\columnwidth}
        \centering
        \includegraphics[width=\linewidth]{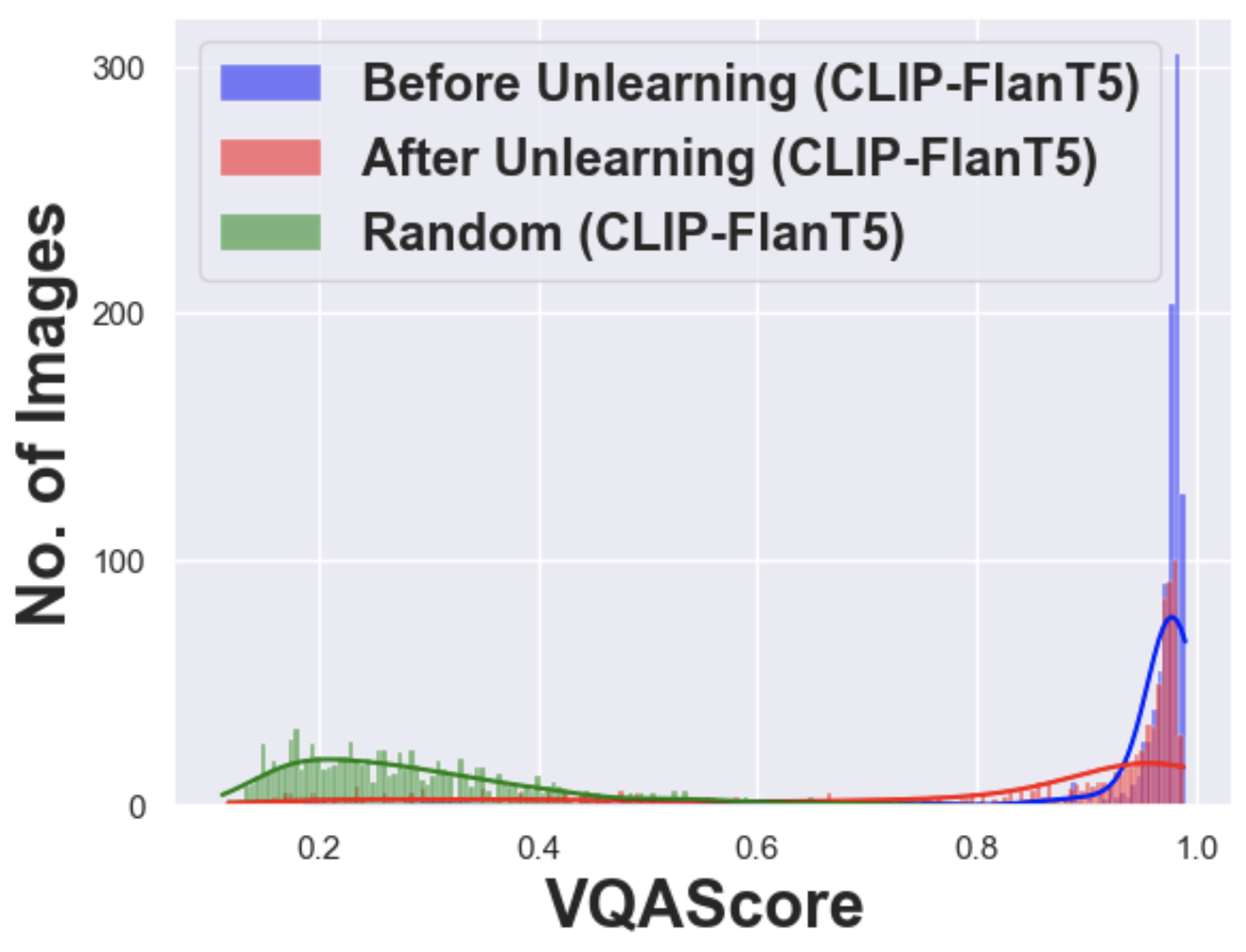}
        \caption{Baseline: Purple Hair}
    \end{subfigure}
    \begin{subfigure}[b]{0.45\columnwidth}
        \centering
        \includegraphics[width=\linewidth]{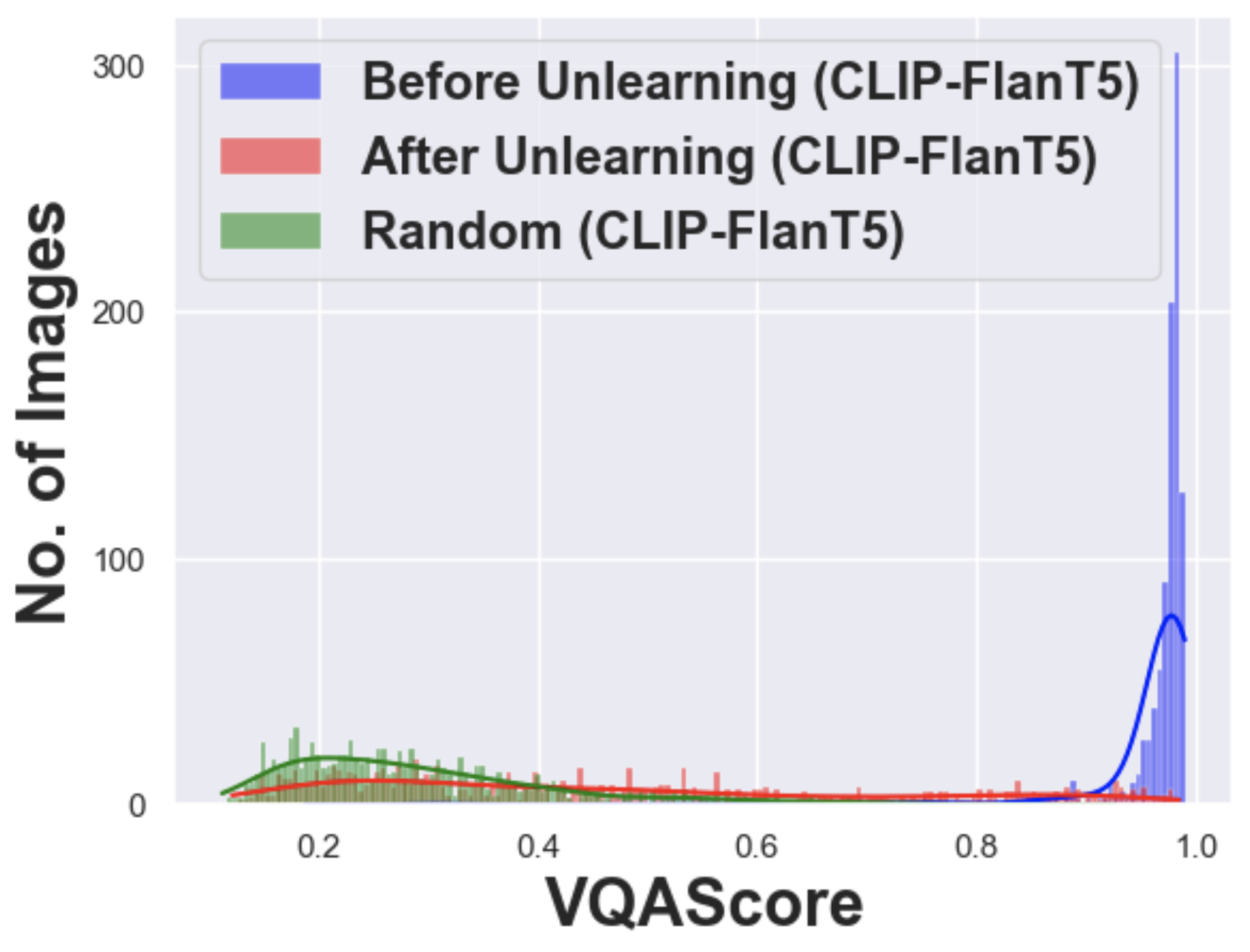}
        \caption{Ours: Purple Hair}
    \end{subfigure}
    \begin{subfigure}[b]{.45\columnwidth}
        \centering
        \includegraphics[width=\linewidth]{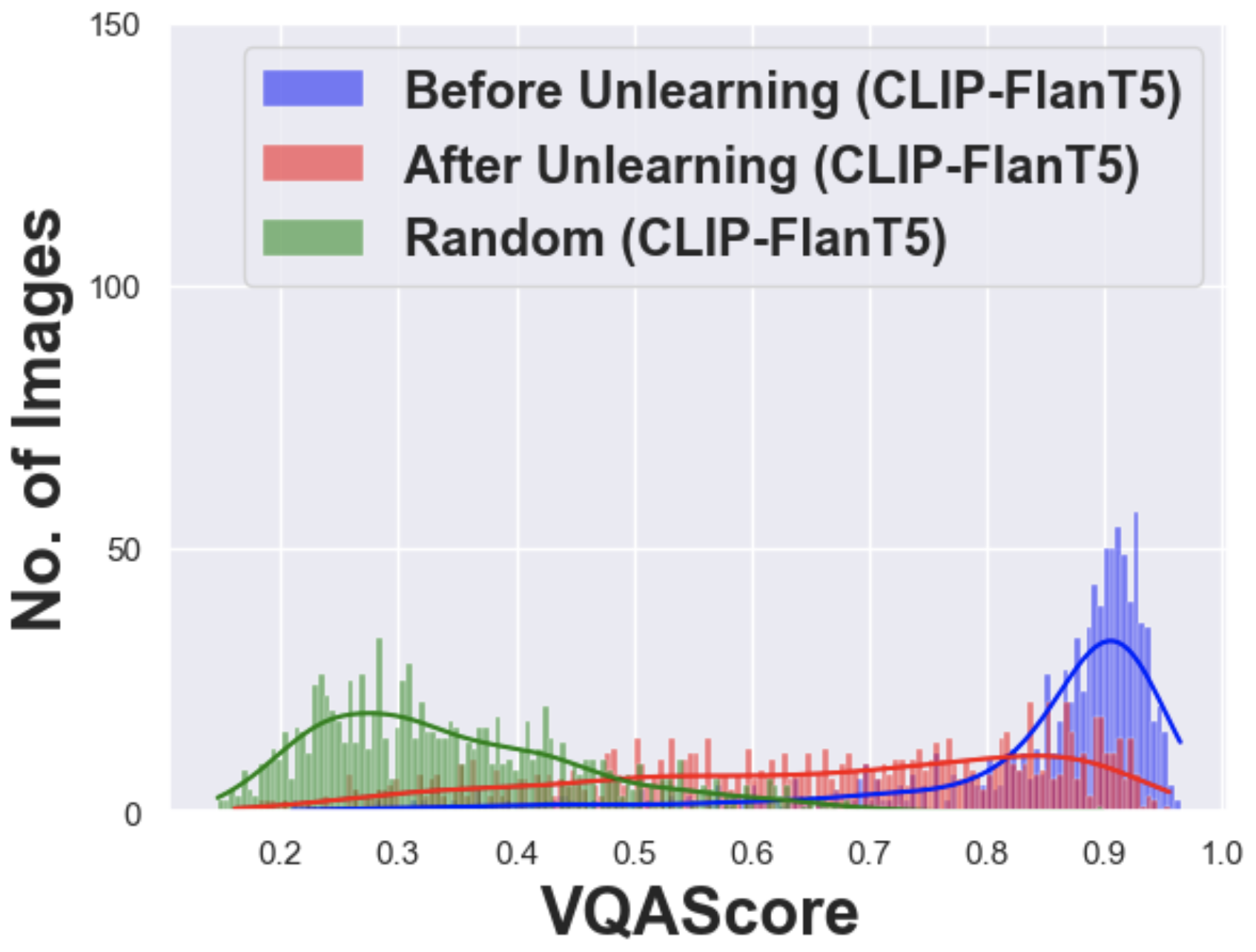}
        \caption{Baseline: Mohawk Hairstyle}
    \end{subfigure}
    \begin{subfigure}[b]{.45\columnwidth}
        \centering
        \includegraphics[width=\linewidth]{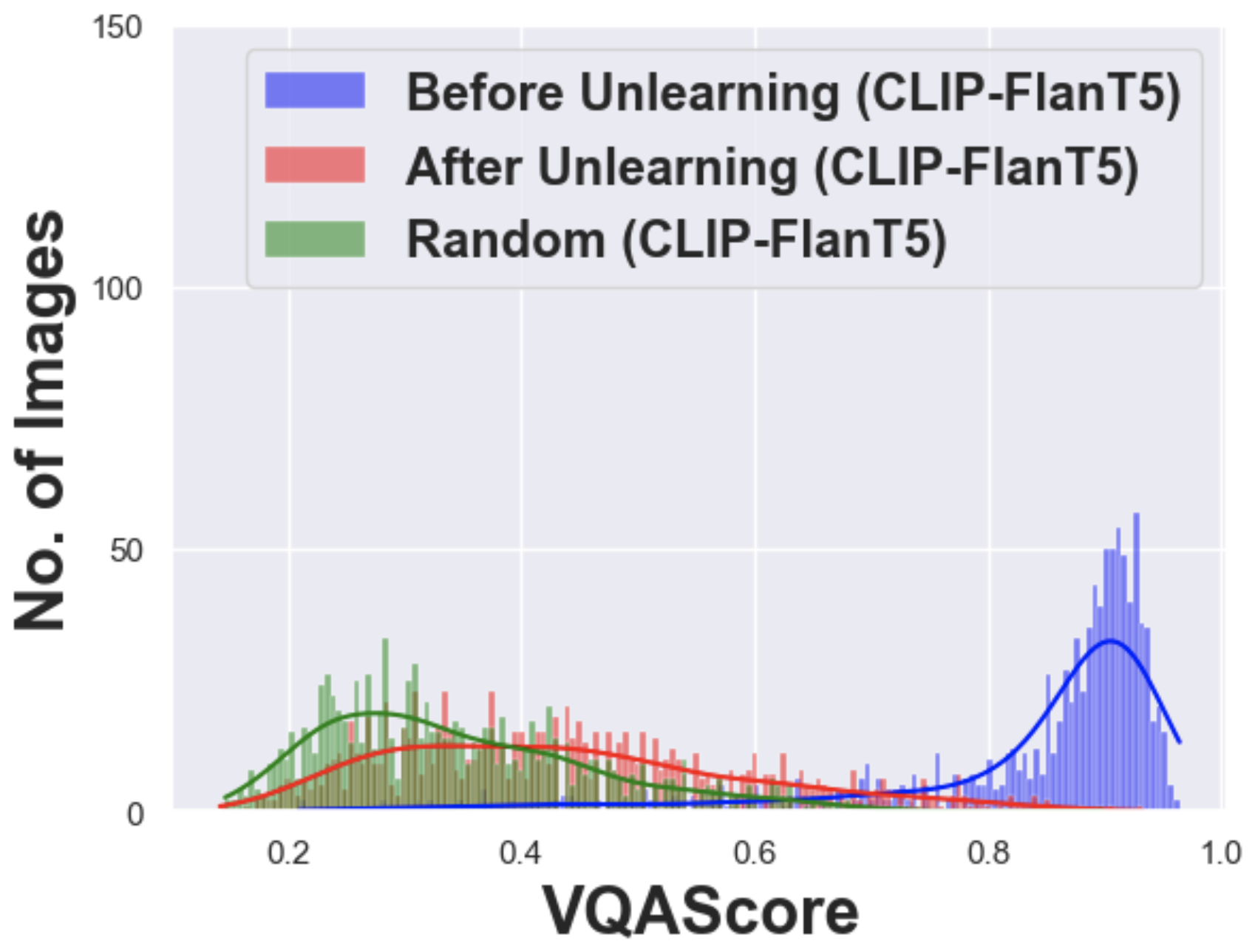}
        \caption{Ours: Mohawk Hairstyle}
    \end{subfigure}
    \caption{Comparison of CLIP-FlanT5 VQAScore distributions for sample text prompts using the baseline method and our \textit{directional unlearning} method.}
    \label{fig:all_dist}
\end{figure}

After unlearning the features, we inspect the usability of the GAN for downstream tasks like StyleCLIP image manipulation. We present some example manipulations using the latent mapper in Figure~\ref{fig:manipulation} after unlearning ``purple hair" (left) and ``spectacles" (right). We see that the GAN cannot generate purple hair even after using a new latent mapper trained on the prompt ``purple hair". However, other edits can be made without training new latent mappers. For example, the Taylor Swift edit in Figure~\ref{fig:manipulation} is identical to the one presented in Figure~\ref{fig:taylor_subfig}. Similarly, after unlearning spectacles, we can still generate edits for an afro hairstyle or purple hair color.
\subsection{Quantitative Evaluation}
We want to quantitatively evaluate unlearning in GANs using our Text-to-Unlearn method, but existing metrics like FID~\cite{fid} and IS~\cite{inc_score} evaluate image fidelity and are not suitable for evaluating unlearning. Sampling latent codes to count undesirable features before and after unlearning~\cite{Moon_2024} is possible but hard to scale for our text-guided approach, requiring classifiers for each prompt. Thus, we focus on creating a scalable and insightful evaluation process. We can formulate this problem as measuring the alignment of the unlearning prompt $p$ with images from the trainable generator $G_{t}$ before and after unlearning. Indeed, the method of measuring this alignment must be capable of capturing the concept in a cross-modal embedding space. 

\noindent \textbf{Evaluation Metrics.} Recent work~\cite{llmscore, lin2024, div_eval} has extensively explored the problem of measuring image-text alignment and moving beyond simple alignment metrics like CLIP score. These new metrics are well-suited to measure the image-text alignment before and after unlearning. We use the image-text matching score (ITM) from the multimodal model BLIP-2~\cite{blip, blip2} and the VQAScore~\cite{lin2024} metric computed using CLIP-FLanT5 XL and LLaVA-1.5 7B~\cite{llava15}. VQAScore has outperformed several image-text alignment baselines and achieved state-of-the-art results. To evaluate identity unlearning, we use a latent mapper to choose latent codes of images that have features of the identity to be unlearned. Then, we compare the images of those latent codes after unlearning using the ArcFace ID network~\cite{arcface}.
\begin{table*}
\centering
\resizebox{\textwidth}{!}{%
\begin{tabular}{|l|llll|llll|llll|}
\hline
\multicolumn{1}{|c|}{\multirow{3}{*}{\begin{tabular}[c]{@{}c@{}}Text\\ Prompt\end{tabular}}} &
  \multicolumn{4}{c|}{CLIP-FlanT5 ($\uparrow$)} &
  \multicolumn{4}{c|}{LLaVA-1.5 ($\uparrow$)} &
  \multicolumn{4}{c|}{BLIP-2 ($\uparrow$)} \\ \cline{2-13} 
\multicolumn{1}{|c|}{} &
  \multicolumn{2}{c|}{In-Domain} &
  \multicolumn{2}{c|}{Out-of-Domain} &
  \multicolumn{2}{c|}{In-Domain} &
  \multicolumn{2}{c|}{Out-of-Domain} &
  \multicolumn{2}{c|}{In-Domain} &
  \multicolumn{2}{c|}{Out-of-Domain} \\ \cline{2-13} 
\multicolumn{1}{|c|}{} &
  \multicolumn{1}{l|}{Baseline} &
  \multicolumn{1}{l|}{Ours} &
  \multicolumn{1}{l|}{Baseline} &
  Ours &
  \multicolumn{1}{l|}{Baseline} &
  \multicolumn{1}{l|}{Ours} &
  \multicolumn{1}{l|}{Baseline} &
  Ours &
  \multicolumn{1}{l|}{Baseline} &
  \multicolumn{1}{l|}{Ours} &
  \multicolumn{1}{l|}{Baseline} &
  Ours \\ \hline
Purple Hair &
  \multicolumn{1}{l|}{0.26} &
  \multicolumn{1}{l|}{\textbf{0.74}} &
  \multicolumn{1}{l|}{0.38} &
  \textbf{0.88} &
  \multicolumn{1}{l|}{0.46} &
  \multicolumn{1}{l|}{\textbf{0.88}} &
  \multicolumn{1}{l|}{0.60} &
  \textbf{0.80} &
  \multicolumn{1}{l|}{0.39} &
  \multicolumn{1}{l|}{\textbf{0.77}} &
  \multicolumn{1}{l|}{0.76} &
  \textbf{0.83} \\ \hline
Mohawk Hairstyle &
  \multicolumn{1}{l|}{0.37} &
  \multicolumn{1}{l|}{\textbf{0.81}} &
  \multicolumn{1}{l|}{0.67} &
  \textbf{0.94} &
  \multicolumn{1}{l|}{0.84} &
  \multicolumn{1}{l|}{\textbf{0.88}} &
  \multicolumn{1}{l|}{0.87} &
  \textbf{0.94} &
  \multicolumn{1}{l|}{0.65} &
  \multicolumn{1}{l|}{\textbf{0.65}} &
  \multicolumn{1}{l|}{0.78} &
  \textbf{0.78} \\ \hline
Spectacles &
  \multicolumn{1}{l|}{0.03} &
  \multicolumn{1}{l|}{\textbf{0.73}} &
  \multicolumn{1}{l|}{0.43} &
  \textbf{0.55} &
  \multicolumn{1}{l|}{0.02} &
  \multicolumn{1}{l|}{\textbf{0.87}} &
  \multicolumn{1}{l|}{0.01} &
  \textbf{0.64} &
  \multicolumn{1}{l|}{0.16} &
  \multicolumn{1}{l|}{\textbf{0.84}} &
  \multicolumn{1}{l|}{0.21} &
  \textbf{0.29} \\ \hline
Curly Long Hair &
  \multicolumn{1}{l|}{0.36} &
  \multicolumn{1}{l|}{\textbf{0.85}} &
  \multicolumn{1}{l|}{0.56} &
  \textbf{0.98} &
  \multicolumn{1}{l|}{0.32} &
  \multicolumn{1}{l|}{\textbf{0.73}} &
  \multicolumn{1}{l|}{0.43} &
  \textbf{0.88} &
  \multicolumn{1}{l|}{0.48} &
  \multicolumn{1}{l|}{\textbf{0.99}} &
  \multicolumn{1}{l|}{0.70} &
  \textbf{0.98} \\ \hline
Surprised &
  \multicolumn{1}{l|}{0.50} &
  \multicolumn{1}{l|}{\textbf{0.76}} &
  \multicolumn{1}{l|}{0.66} &
  \textbf{0.72} &
  \multicolumn{1}{l|}{0.31} &
  \multicolumn{1}{l|}{\textbf{0.70}} &
  \multicolumn{1}{l|}{0.46} &
  \textbf{0.73} &
  \multicolumn{1}{l|}{0.42} &
  \multicolumn{1}{l|}{\textbf{0.78}} &
  \multicolumn{1}{l|}{0.62} &
  \textbf{0.95} \\ \hline
Angry &
  \multicolumn{1}{l|}{0.10} &
  \multicolumn{1}{l|}{\textbf{0.62}} &
  \multicolumn{1}{l|}{0.20} &
  \textbf{0.82} &
  \multicolumn{1}{l|}{0.16} &
  \multicolumn{1}{l|}{\textbf{0.84}} &
  \multicolumn{1}{l|}{0.25} &
  \textbf{0.92} &
  \multicolumn{1}{l|}{0.17} &
  \multicolumn{1}{l|}{\textbf{0.81}} &
  \multicolumn{1}{l|}{0.25} &
  \textbf{0.96} \\ \hline
Afro Hairstyle &
  \multicolumn{1}{l|}{0.62} &
  \multicolumn{1}{l|}{\textbf{0.89}} &
  \multicolumn{1}{l|}{0.65} &
  \textbf{0.82} &
  \multicolumn{1}{l|}{0.59} &
  \multicolumn{1}{l|}{\textbf{0.96}} &
  \multicolumn{1}{l|}{0.68} &
  \textbf{0.89} &
  \multicolumn{1}{l|}{0.51} &
  \multicolumn{1}{l|}{\textbf{0.99}} &
  \multicolumn{1}{l|}{0.74} &
  \textbf{0.94} \\ \hline
Makeup &
  \multicolumn{1}{l|}{0.14} &
  \multicolumn{1}{l|}{\textbf{0.89}} &
  \multicolumn{1}{l|}{0.26} &
  \textbf{0.99} &
  \multicolumn{1}{l|}{0.12} &
  \multicolumn{1}{l|}{\textbf{0.86}} &
  \multicolumn{1}{l|}{0.21} &
  \textbf{0.97} &
  \multicolumn{1}{l|}{0.18} &
  \multicolumn{1}{l|}{\textbf{0.51}} &
  \multicolumn{1}{l|}{0.60} &
  \textbf{0.62} \\ \hline
Bobcut Hairstyle &
  \multicolumn{1}{l|}{0.69} &
  \multicolumn{1}{l|}{\textbf{0.70}} &
  \multicolumn{1}{l|}{0.59} &
  \textbf{0.80} &
  \multicolumn{1}{l|}{0.35} &
  \multicolumn{1}{l|}{\textbf{0.39}} &
  \multicolumn{1}{l|}{0.35} &
  \textbf{0.56} &
  \multicolumn{1}{l|}{0.36} &
  \multicolumn{1}{l|}{\textbf{0.40}} &
  \multicolumn{1}{l|}{0.57} &
  \textbf{0.66} \\ \hline
\end{tabular}%
}
\caption{Degree of unlearning ($\gamma$) computed using various image-text alignment scoring metrics for \textit{in-domain} and \textit{out-of-domain} images. Higher scores are better ($\uparrow$) and are highlighted in bold.}
\label{tab:dou}
\end{table*}

\noindent \textbf{Baseline.}
Since there is no relevant work that uses only text to unlearn from GANs, we employ an intuitive baseline: We use the latent mapper to generate negative samples (images containing the feature or identity to be unlearned) from $G_{t}$ and simply maximize the CLIP loss with respect to the unlearning prompt, \ie, maximize $\mathcal{L}_{CLIP}(\hat{x_{t}}, p)$. $\hat{x_{t}}$ is the synthesized image during training and $p$ is the prompt. This approach does not use the directional loss from our method. For example, while unlearning purple hair, we would maximize the loss of each image during unlearning against the text prompt ``purple hair" via CLIP.

\noindent \textbf{Evaluation Method.} 
For each text prompt, we use a latent mapper to help sample 1000 images from the GAN (\textit{in-domain} images) before and after unlearning. Initially, most of the samples generated will have the undesired attribute, but post-unlearning, most will not. Then, we compute CLIP-FlanT5 VQAScore, LLaVA VQAScore, and BLIP-2 ITM score distributions for both sets of samples. Additionally, for each prompt, we compute the image-text score distribution on 1000 randomly sampled images as a reference. After unlearning, the score distribution should be similar to the random score distribution. We use this reference because an image-text pair often has a non-zero CLIP score even if the prompt is completely unrelated to the image.
Some example plots are shown in Figure~\ref{fig:all_dist}.
Our objective is to maximize the ``distance" between the blue histogram (before unlearning) and the red histogram (after unlearning). Thus, we propose our metric, \emph{degree of unlearning ($\gamma$)} in Equation~\ref{eq:dou}: 
\begin{equation}\label{eq:dou}
    \gamma = \frac{W_1(A,B)}{W_1(B,R)}
\end{equation}
$W_1(\cdot,\cdot)$ is the Wasserstein 1-distance between
two distributions. $A$, $B$, and $R$ are score distributions after unlearning, before unlearning, and for the random images, respectively. By score distribution, we refer to the individual image-text score distribution obtained using either CLIP-FlanT5, LLaVA, or BLIP-2. We use the Wasserstein 1-distance because it compares the histograms without making assumptions about the underlying distribution and is suitable for ordered data.

Besides the \textit{in-domain} evaluation, we assess our unlearning method on \textit{out-of-domain} data by encoding 1000 CelebAHQ images into the GAN’s latent space using the e4e encoder. We then calculate the same score distributions to confirm that the unlearned model generalizes effectively to these images, ensuring its reliability for downstream tasks like image editing (shown in Table~\ref{tab:dou}).

Clearly, directional unlearning outperforms the baseline for all text prompts. 
In Figure~\ref{fig:all_dist}, we see that the blue and red histograms are much more separated using our method as opposed to the baseline method. We include the average ID scores for identity unlearning in Table~\ref{tab:id_tab} comparing our method against the baseline method. For each identity we considered, our method outperformed the baseline method. We refer the readers to the supplementary material (Section~\ref{sec:stability}) for a detailed analysis showing the stability of our unlearning strategy.
\begin{table}
\centering
\resizebox{\columnwidth}{!}{%
\begin{tabular}{|l|l|l|l|}
\hline
Prompt        & Taylor Swift & Donald Trump & Tupac Shakur \\ \hline
ID (Baseline) $\downarrow$ & 0.38         & 0.82         & 0.88         \\ \hline
ID (Ours) $\downarrow$     & \textbf{0.2}          & \textbf{0.3}          & \textbf{0.5}          \\ \hline
\end{tabular}%
}
\caption{ID scores for the baseline method and our method after unlearning different identities computed using 5000 samples. Lower scores are better ($\downarrow$).}
\label{tab:id_tab}
\end{table}

Apart from quantifying the degree of unlearning, we evaluate the extent to which our unlearning method affects image generation of other features. First, we sample 400 images per feature for a set of four features (``purple hair", ``spectacles", ``surprised", ``afro hairstyle") from the GAN prior to unlearning and compute the average VQAScore for each prompt as a baseline. Then, we unlearn each feature and evaluate the change in the mean VQAScore for the other three features. In Table~\ref{tab:specificity}, we report the shift in mean VQAScores from the baseline. We see that there is marginal shift in the scores for unrelated features, suggesting that our method supports disentangled unlearning.
\begin{table}
\centering
\resizebox{\columnwidth}{!}{%
\begin{tabular}{|l|l|l|l|l|}
\hline
Feature        & Purple Hair & Spectacles & Surprised & Afro Hairstyle \\ \hline
Purple Hair    & -60\%       &   +1.2\%   &  -0.2\%   &  -0.2\%        \\ \hline
Spectacles     & -0.4\%      &  -30.4\%   &  +1\%     &  -1\%          \\ \hline
Surprised      & -0.4\%      &   +1\%     &  -20\%    &  -1.1\%        \\ \hline
Afro Hairstyle & -0.7\%      &   +0.7\%   &   +0.8\%  &  -44.8\%       \\ \hline
\end{tabular}%
}
\caption{Quantitative results for the effect of unlearning each feature (rows) on the VQAScore of other unrelated features (columns). Each entry is a percentage change of the CLIP-FlanT5 VQAScore for that feature with respect to its baseline score before unlearning.}
\label{tab:specificity}
\end{table}

\begin{figure}
    \centering
    \begin{subfigure}[b]{0.33\columnwidth}
        \centering
        \includegraphics[width=\linewidth]{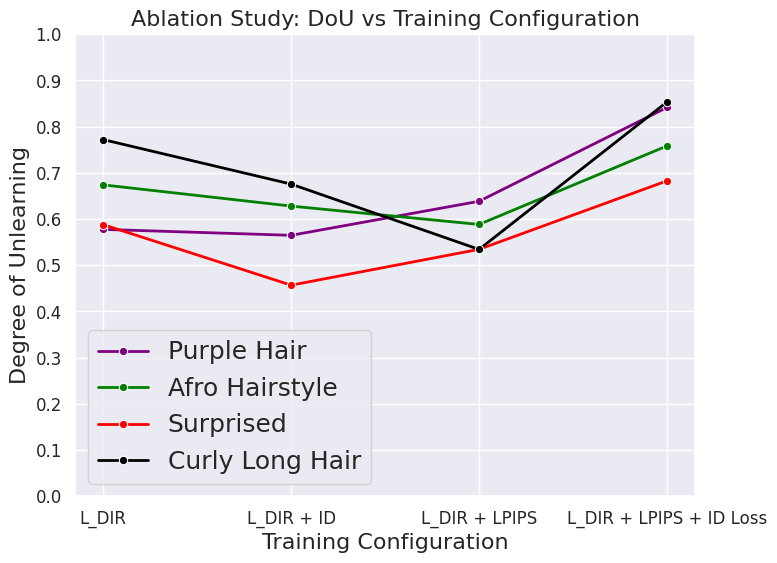}
        \caption{Ablation on loss components (directional loss, ID loss, and LPIPS loss) of $\mathcal{L}_{u}$.}
        \label{fig:loss_ablation}
    \end{subfigure}
    \begin{subfigure}[b]{0.3\columnwidth}
        \centering
        \includegraphics[width=\linewidth]{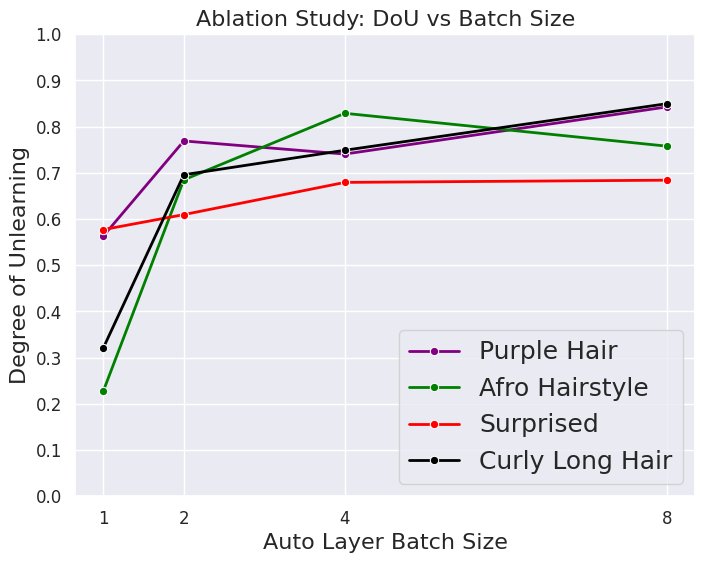}
        \caption{Ablation on batch size used for computing the reference direction $\vec{i}$ in Phase 1.}
        \label{fig:img_ref_ablation}
    \end{subfigure}
    \begin{subfigure}[b]{0.3\columnwidth}
        \centering
        \includegraphics[width=\linewidth]{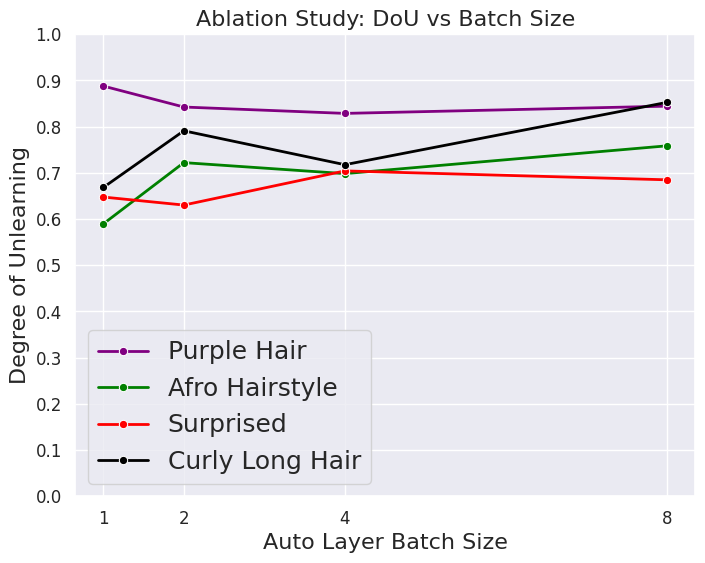}
        \caption{Ablation on batch size used for automatic layer selection in Phase 2.}
        \label{fig:auto_layer_ablation}
    \end{subfigure}
    \caption{Ablation experiments for relevant hyperparameters.}
    \label{fig:ablation}
\end{figure}

\noindent \textbf{Ablation Study.} We perform three ablation experiments (as shown in Figure~\ref{fig:ablation}): impact of (i) loss function components, (ii) batch size of the automatic layer selection strategy in Phase 2, and (iii) batch size used when computing the reference direction $\vec{i}$ (in Phase 1) on the degree of unlearning.  
The ideal batch size, for the automatic layer selection strategy and for computing the reference direction $\vec{i}$ is 8 based on the stability across all prompts. We also see that both the LPIPS loss and ID loss are needed for maximal unlearning.
\section{Limitations, Conclusion, and Future Work}
In this paper, we propose Text-to-Unlearn, a method to unlearn concepts from a GAN using only text prompts. Our experiments show that Text-to-Unlearn can achieve favorable results at different levels of granularity and we validate this using our metric: degree of unlearning ($\gamma$). It is important to acknowledge that our method relies on a pre-trained CLIP model to guide the unlearning process, and thus, text prompts that are not well-represented by CLIP's visual encoder cannot be expected to achieve effective unlearning. Furthermore, pre-trained VLMs like CLIP are known to contain harmful societal biases 
and these can adversely influence the unlearning procedure. Recent work by~\citeauthor{saner} and~\citeauthor{debias} propose ways to debias pre-trained VLMs, which we plan to incorporate in our future work.
{
    \small
    \bibliographystyle{ieeenat_fullname}
    \bibliography{main}
}

\clearpage
\setcounter{page}{1}
\maketitlesupplementary

\section{Additional VQAScore Distributions}
\begin{figure}
    \centering
    \begin{subfigure}[b]{0.3\columnwidth}
        \centering
        \includegraphics[width=\linewidth]{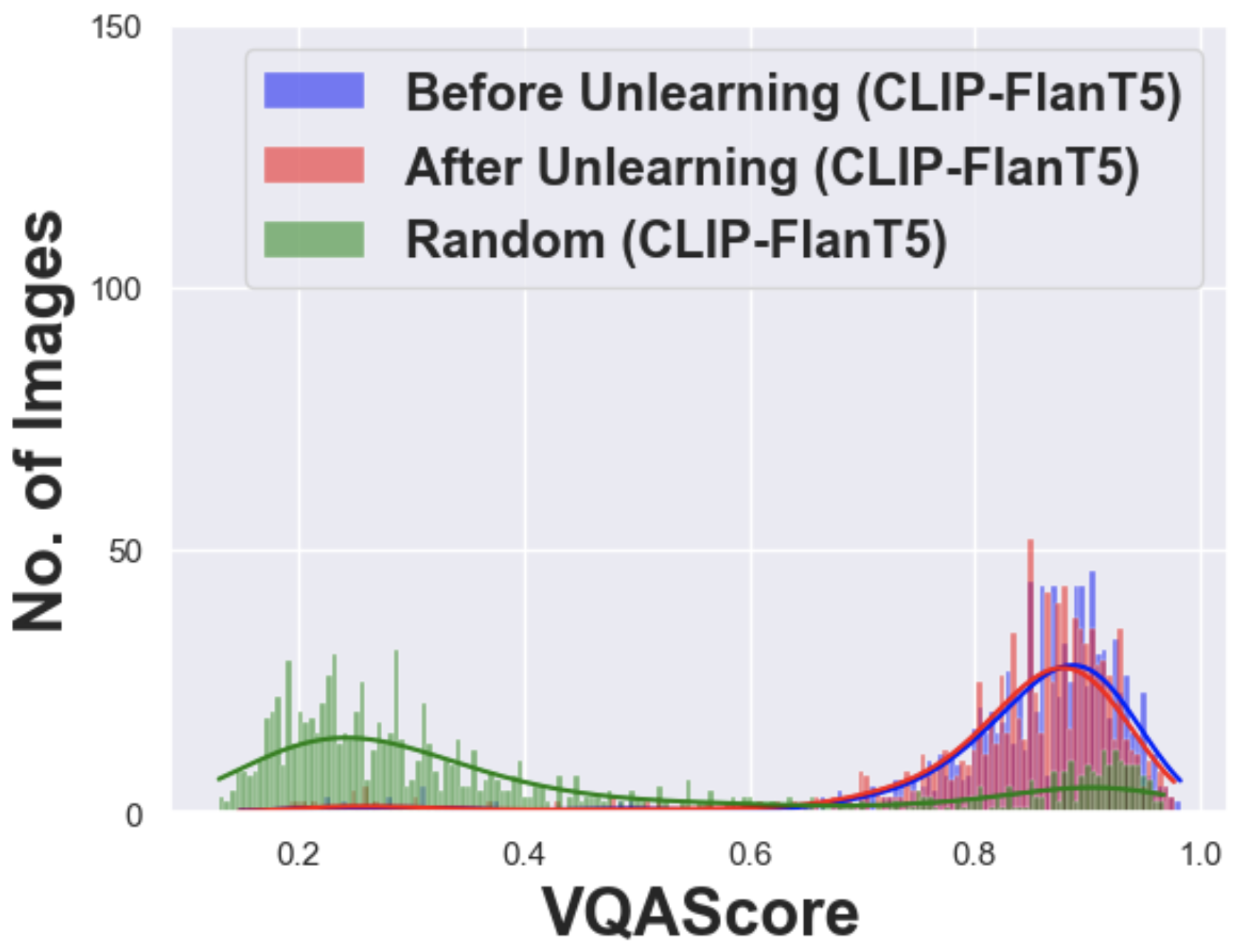}
        \caption{Spectacles}
        \label{fig:t5_spectacles_b}
    \end{subfigure}
    \begin{subfigure}[b]{0.3\columnwidth}
        \centering
        \includegraphics[width=\linewidth]{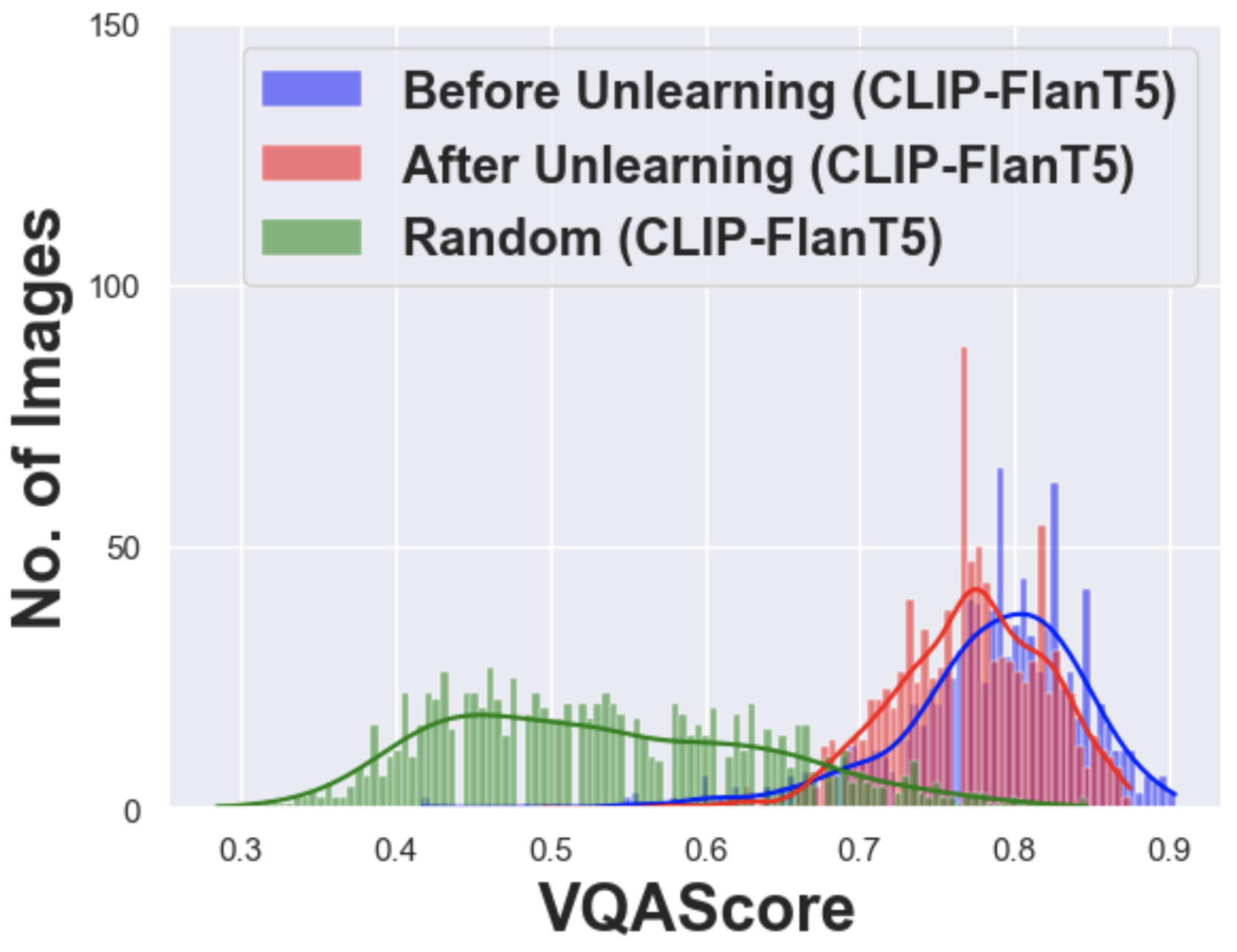}
        \caption{Surprised}
        \label{fig:t5_surprised_b}
    \end{subfigure}
    \begin{subfigure}[b]{0.32\columnwidth}
        \centering
        \includegraphics[width=\linewidth]{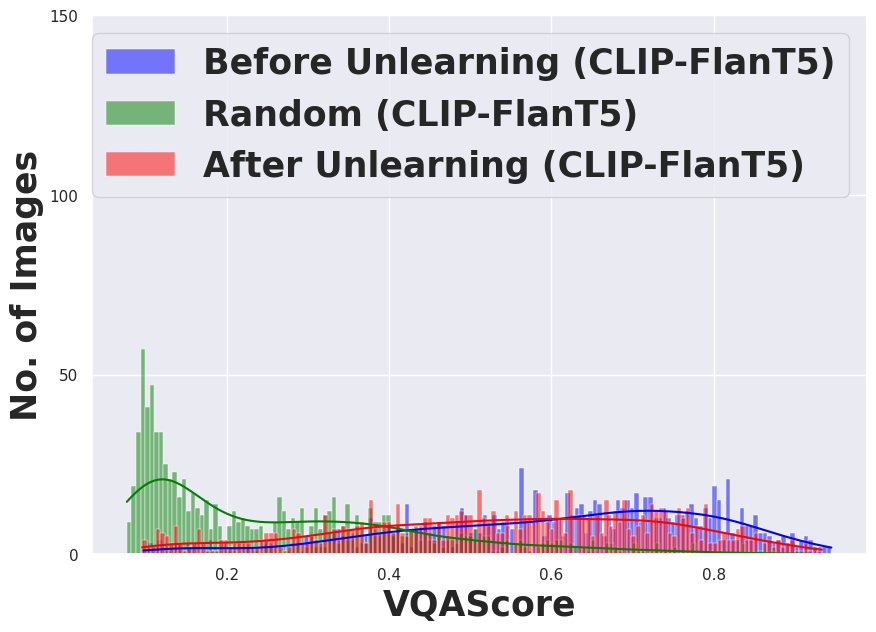}
        \caption{Angry}
        \label{fig:t5_angry_b}
    \end{subfigure}
    \caption{CLIP-FlanT5 VQAScore distribution computed over 1000 images before and after unlearning for different text prompts using the baseline method.}
    \label{fig:add_vqa_b}
\end{figure}

\begin{figure}
    \centering
    \begin{subfigure}[b]{0.3\columnwidth}
        \centering
        \includegraphics[width=\linewidth]{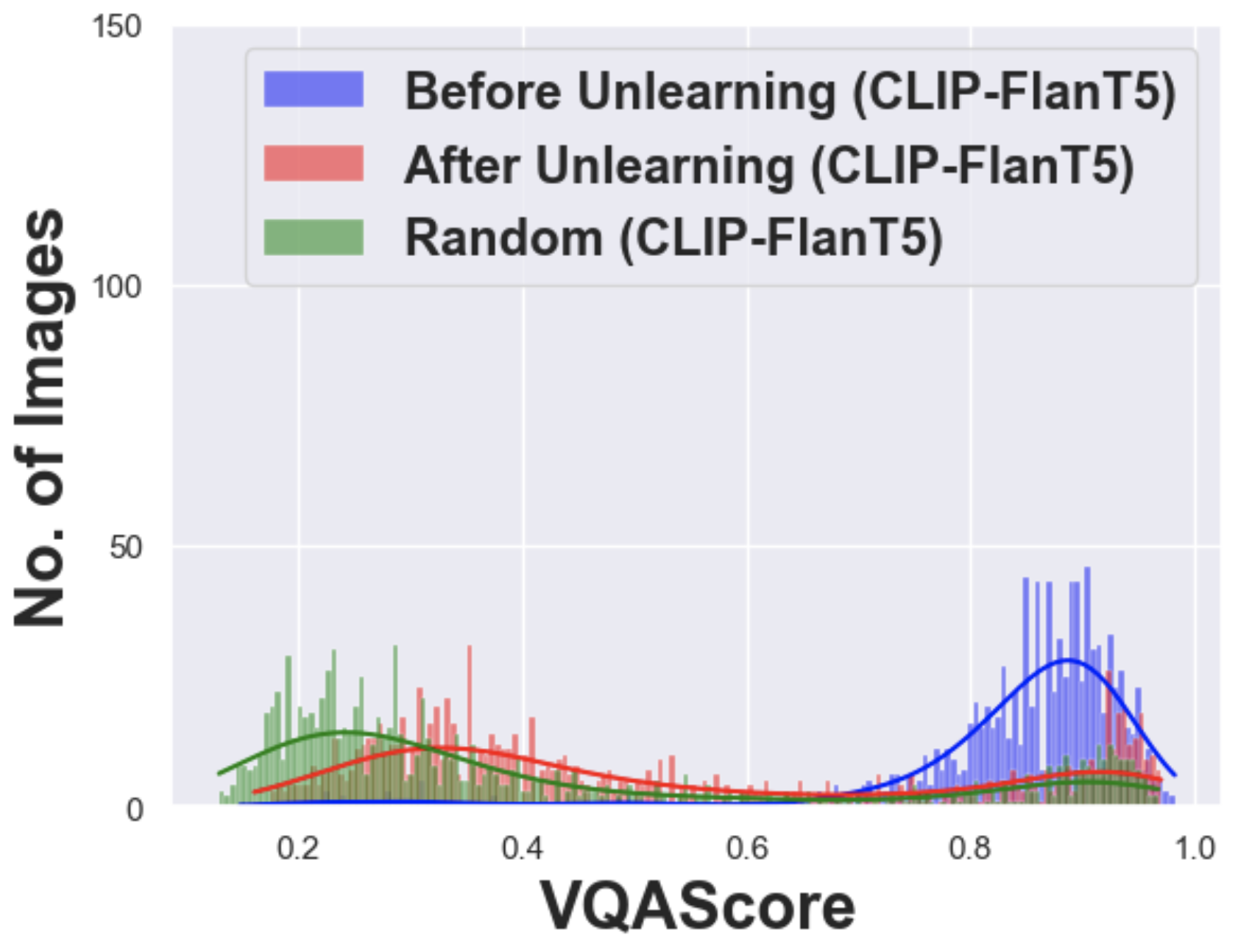}
        \caption{Spectacles}
        \label{fig:t5_spectacles}
    \end{subfigure}
    \begin{subfigure}[b]{0.3\columnwidth}
        \centering
        \includegraphics[width=\linewidth]{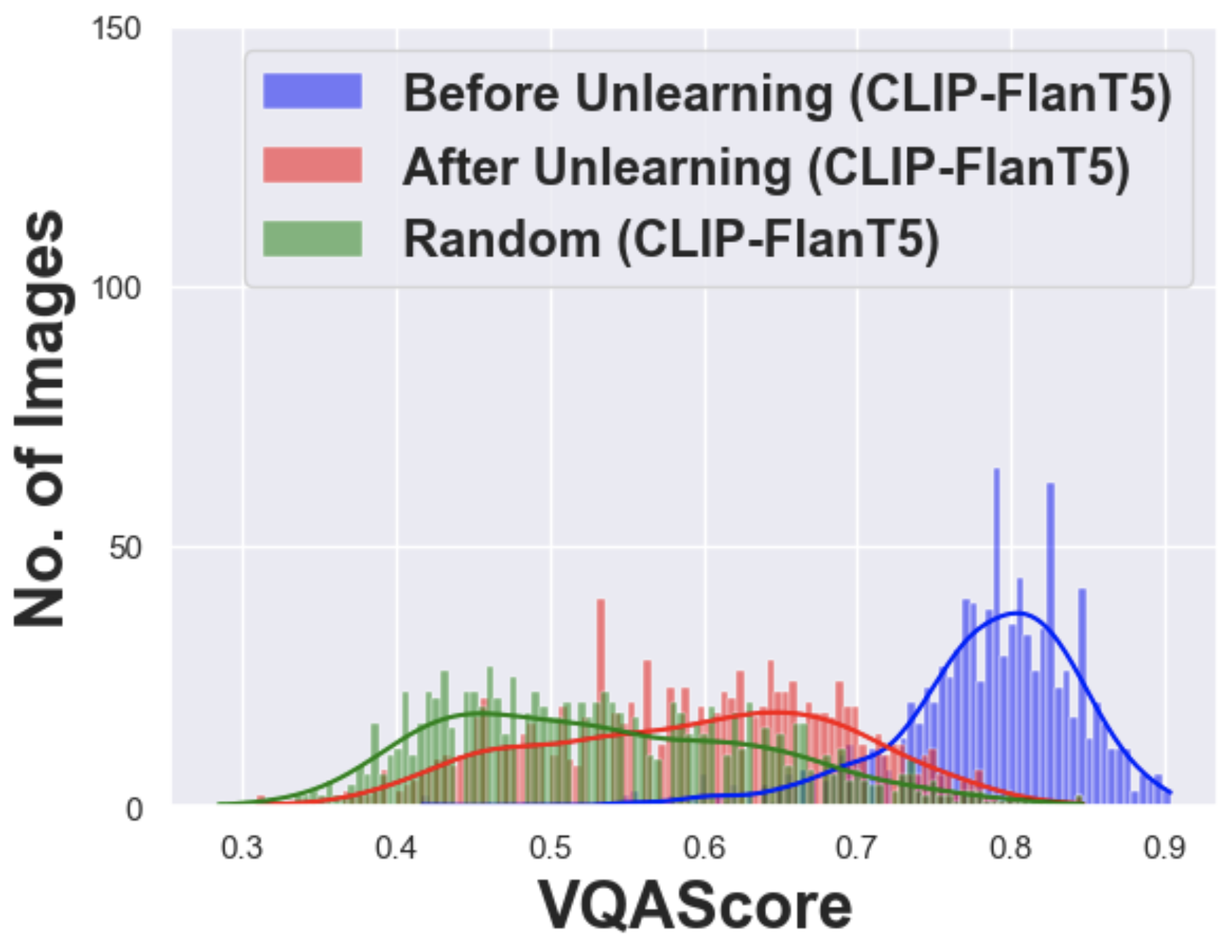}
        \caption{Surprised}
        \label{fig:t5_surprised}
    \end{subfigure}
    \begin{subfigure}[b]{0.32\columnwidth}
        \centering
        \includegraphics[width=\linewidth]{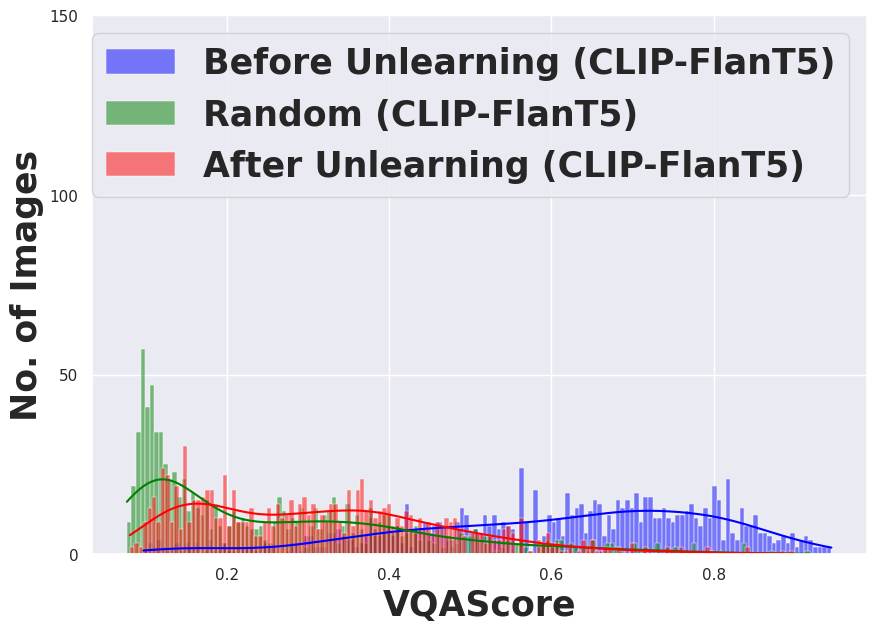}
        \caption{Angry}
        \label{fig:t5_angry}
    \end{subfigure}
    \caption{CLIP-FlanT5 VQAScore distribution computed over 1000 images before and after unlearning for different text prompts using our \textit{directional unlearning} method.}
    \label{fig:add_vqa}
\end{figure}

We include additional graphs showing the VQAScore distributions for unlearning some other text prompts in Figures~\ref{fig:add_vqa_b} and~\ref{fig:add_vqa}. Clearly, the red and blue distributions are much more separated using our method.

\section{Details on VQAScore Metric}
In this section, we elaborate on the VQAScore image-text alignment metric used in Section~\ref{sec:experiments}. VQA models are designed to answer questions about images and we evaluate the image-text alignment by querying the model with the question ``Does this figure show \{\emph{text}\}? Please answer yes or no." The VQAScore presented by ~\citeauthor{lin2024} is computed as the probability that the answer is yes given a question and image, i.e., P(``Yes" $|$ question, image). Despite being simplistic, it has been shown to outperform several image-text alignment baselines and achieve SOTA results.

\section{Training Details}
Here, we provide detailed instructions and hyperparameters used for training the latent mapper and $G_{t}$ for various unlearning tasks. Unlike StyleCLIP, we train the latent mapper on samples from the latent space of the GAN since we do not use external datasets. Our hyperparameters are different from StyleCLIP for certain prompts.

There are 3 hyperparameters for the latent mapper training: (i) ID loss regularization ($\lambda_{ID}$), (ii) L2 loss regularization ($\lambda_{L2}$), and (iii) Step magnitude in the $\mathcal{W^+}$ space ($\delta$). In practice, the latent mapper is implemented as $\hat{w} = w + \delta \cdot M_p(w)$ to ensure gradients are updated stably. The training parameters are listed below:
\begin{table}[ht]
\centering
\resizebox{\columnwidth}{!}{%
\begin{tabular}{|l|l|l|l|l|}
\hline
Text Prompt      & $\lambda_{ID}$  & $\lambda_{L2}$  & $\delta$ & Levels               \\ \hline
Purple Hair      & 0.1             & 0.8            & 0.1   & fine, medium, coarse \\
Mohawk Hairstyle & 0.1             & 0.8            & 0.8   & medium, coarse       \\
Spectacles       & 0.1             & 0.8            & 0.9   & medium, coarse       \\
Curly Long Hair  & 0.1             & 0.8            & 0.8   & medium, coarse       \\
Surprised        & 0.1             & 0.8            & 0.5   & medium, coarse, fine       \\
Angry            & 0.1             & 0.8            & 0.3   & medium, coarse, fine       \\
Afro Hairstyle   & 0.1             & 0.8            & 0.8   & medium, coarse       \\
Makeup           & 0.1             & 0.8            & 0.3   & medium, coarse, fine       \\
Bobcut Hairstyle & 0.1             & 0.8            & 0.3   & medium, coarse       \\
Taylor Swift     & 0               & 0.8            & 0.1   & fine, medium, coarse \\
Donald Trump     & 0               & 1.5            & 0.1   & fine, medium, coarse \\
Tupac Shakur     & 0               & 1.5            & 0.1   & fine, medium, coarse \\ \hline
\end{tabular}%
}
\caption{Hyperparameters for training the latent mapper.}
\label{tab:mapper_params}
\end{table}

Additionally, in Table~\ref{tab:mapper_params}, we include the architecture of the multi-level mapper used for each text prompt. The levels correspond to the same scheme presented in StyleCLIP. As a rule of thumb, if no change in color is required, we omit the finegrained level from the mapper. As such, identities will require all levels enabled.

For the directional unlearning procedure, we have three hyperparameters: (i) Learning Rate ($lr$), (ii) ID loss regularization ($\lambda_{ID}$), and (iii) LPIPS loss regularization ($\lambda_{lpips}$). The hyperparameters to reproduce our results are:
\begin{table}[ht]
\centering
\resizebox{0.75\columnwidth}{!}{%
\begin{tabular}{|l|l|l|l|}
\hline
Text Prompt      & $lr$   & $\lambda_{ID}$   & $\lambda_{lpips}$ \\ \hline
Purple Hair      & 8e-3 & 4e-3 & 1e-1  \\
Mohawk Hairstyle & 8e-3 & 4e-3 & 1e-1  \\
Spectacles       & 1e-2 & 2e-3 & 1e-1  \\
Curly Long Hair  & 8e-3 & 4e-3 & 1e-1  \\
Surprised        & 8e-3 & 4e-3 & 1e-1  \\
Angry            & 8e-3 & 4e-3 & 1e-1  \\
Afro Hairstyle   & 8e-3 & 4e-3 & 1e-1  \\
Makeup           & 8e-3 & 4e-3 & 1e-1  \\
Bobcut Hairstyle & 8e-3 & 4e-3 & 1e-1  \\
Taylor Swift     & 8e-3 & 0    & 1e-1  \\
Donald Trump     & 8e-3 & 0    & 1e-1  \\
Tupac Shakur     & 8e-3 & 0    & 1e-1  \\ \hline
\end{tabular}%
}
\caption{Hyperparameters for unlearning.}
\label{tab:unlearning_params}
\end{table}

In Table~\ref{tab:unlearning_params}, $\lambda_{ID}$ is 0 since this is not a loss component for identity unlearning as discussed in the main paper.

\section{Discussion on Training Stability}
\label{sec:stability}
\begin{figure}[ht]
    \centering
    \includegraphics[width=0.8\linewidth]{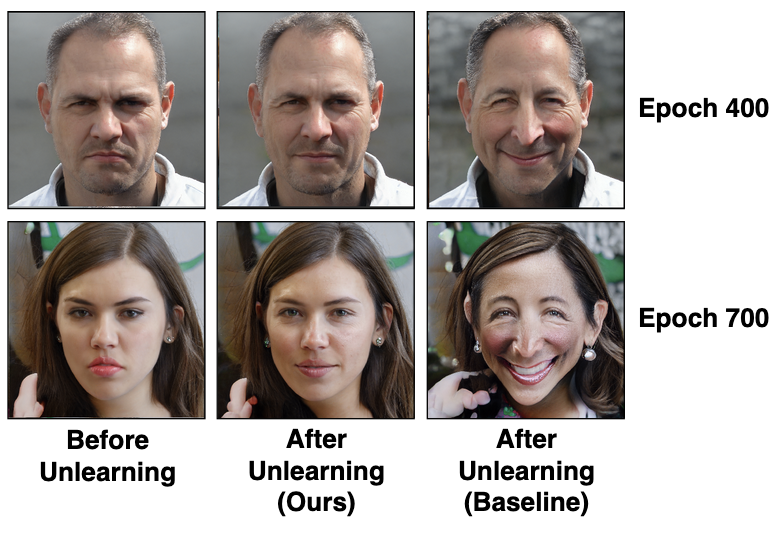}
    \caption{Qualitative comparison between directional unlearning (ours) and baseline method for the prompt ``angry". Left most image was generated using a latent mapper trained on ``angry".}
    \label{fig:stability}
\end{figure}

Here, we discuss the stability provided by \emph{directional unlearning} during the unlearning process. Based on Figure~\ref{fig:all_dist}, one could think of increasing the learning rate for the baseline method to achieve better unlearning. Figure~\ref{fig:stability} shows the results of unlearning after 400 and 700 steps. Using our \emph{directional unlearning} method, we can subtly unlearn the ``angry" expression whereas the baseline method causes distortion in the images generated. Furthermore, as we continue to fine-tune for a longer number of steps, the quality of images will not reduce because we unlearn only along a precomputed direction (from Equation~\ref{eq:ref_direction}).

After unlearning for 800 steps, the FID (lower scores represent higher fidelity) using our method was 6.98 as opposed to 49.1 from the baseline method. The FID was computed using 10000 samples for each of the unlearned models. However, lower learning rates using the baseline method can avoid distortion but achieve little to none unlearning as seen in Figure~\ref{fig:add_vqa_b}.

\section{System and Hardware Details}
All our code was tested on Ubuntu 22.04 with PyTorch 2.1. In terms of hardware requirements, the latent mapper and unlearning can be implemented using any GPU architecture. The latent mapper training can be done on a T4 GPU. The unlearning requires at least 24GB of GPU RAM and thus, we implemented this on an NVIDIA A10G. However, this should work the same on an NVIDIA 3090. The evaluation scripts can only be run on a GPU with NVIDIA Ampere architecture (\eg, A10G, A100, \etc). GPUs like V100 do not support the VQAScore method due to a dependency on the t2v-metrics library. It may be possible if built from source and the dependency on the bfloat data type is removed, however, we have not tested this.

\section{Prompt Engineering during Evaluation}
During evaluation, the ``surprised" feature was evaluated with the text caption ``surprised with mouth open" since the surprised edit using the latent mapper generates images of faces with their mouth open. Unlike CLIP's text encoder, the VQA models can capture the image-text alignment better with a more detailed prompt. We suggest using this approach when evaluating other fine-grained edits as the objective is not to evaluate the VQA model, but to evaluate the image-text alignment before and after unlearning. All other prompts in the paper were evaluated with the same captions used for unlearning (\eg, ``purple hair", \etc)

\section{Algorithms}
We briefly outline the unlearning algorithm for feature unlearning in Algorithm~\ref{alg:unlearn_alg1}.
\begin{algorithm}
\caption{Feature Unlearning using our Directional Unlearning Method}\label{alg:unlearn_alg1}
\begin{algorithmic}
\Require Mapper ($M_p$), $G_{t}$, $G_{f}$, prompt ($p$), step size ($\delta$), total steps ($s_{max}$), batch size ($b$)\\
\State $z \gets \mathcal{N}^{8 \times 512}(0,1)$
\State $i \gets \text{compute\_ref\_direction}(z)$ 
\Comment Phase 1
\State $s \gets 0$
\While{$s < s_{max}$} \Comment{Phase 2}
    \State $layer\_selection(G_{t}, p)$
    \State $z \gets \mathcal{N}^{b \times 512}(0,1)$
    \State $w \gets G_{t}.\text{map}(z)$
    \State $\hat{w} \gets w + \delta M_p(w)$
    \State $\hat{x_{f}} \gets G_{f}.\text{synthesis}(\hat{w})$
    \State $\hat{x_{t}} \gets G_{t}.\text{synthesis}(\hat{w})$
    \State compute\_loss($\hat{x_{t}},~\hat{x_{f}},~i$)
    \State update $G_{t}$
    \State $s \gets s + 1$
\EndWhile
\end{algorithmic}
\end{algorithm}

 The pseudo-code for identity unlearning is detailed in Algorithm~\ref{alg:unlearn_alg2}.
\begin{algorithm}
\caption{Identity Unlearning using our Directional Unlearning Method}\label{alg:unlearn_alg2}
\begin{algorithmic}
\Require Mapper ($M_p$), $G_{t}$, $G_{f}$, prompt ($p$), step size ($\delta$), total steps ($s_{max}$), batch size ($b$)\\
\State $z \gets \mathcal{N}^{8 \times 512}(0,1)$
\State $\overline{w} \gets G_{t}.\text{mean\_latent()}$
\State $i \gets \text{compute\_ref\_direction}(z, \overline{w})$ \Comment Phase 1
\State $s \gets 0$
\While{$s < s_{max}$} \Comment{Phase 2}
    \State $layer\_selection(G_{t}, p)$
    \State $z \gets \mathcal{N}^{b \times 512}(0,1)$
    \State $w \gets G_{t}.\text{map}(z)$
    \State $\hat{w} \gets w + \delta M_p(w)$
    \State $\hat{x_{f}} \gets G_{f}.\text{synthesis}(\hat{w})$
    \State $\hat{x_{t}} \gets G_{t}.\text{synthesis}(\hat{w})$
    \State compute\_loss($\hat{x_{t}},~\hat{x_{f}},~\overline{w}$) \Comment direct toward mean face
    \State update $G_{t}$
    \State $s \gets s + 1$
\EndWhile
\end{algorithmic}
\end{algorithm}

Algorithm~\ref{alg:unlearn_alg3} is the baseline algorithm used in this paper.
\begin{algorithm}
\caption{Baseline Unlearning Algorithm}\label{alg:unlearn_alg3}
\begin{algorithmic}
\Require Mapper ($M_p$), $G_{t}$, prompt ($p$), step size ($\delta$), total steps ($s_{max}$), batch size ($b$)\\
\State $z \gets \mathcal{N}^{8 \times 512}(0,1)$ 
\State $s \gets 0$
\While{$s < s_{max}$} 
    \State $layer\_selection(G_{t}, p)$
    \State $z \gets \mathcal{N}^{b \times 512}(0,1)$
    \State $w \gets G_{t}.\text{map}(z)$
    \State $\hat{w} \gets w + \delta M_p(w)$
    \State $\hat{x_{t}} \gets G_{t}.\text{synthesis}(\hat{w})$
    \State clip\_loss($\hat{x_{t}},p$) \Comment Global CLIP loss
    \State update $G_{t}$
    \State $s \gets s + 1$
\EndWhile
\end{algorithmic}
\end{algorithm}

\end{document}